\documentclass{article}
\usepackage{ijcai19}
\usepackage[square,sort,comma,semicolon,numbers,authoryear]{natbib}

\usepackage{times}
\usepackage{dblfloatfix}
\usepackage{soul}
\usepackage{url}
\usepackage[utf8]{inputenc}
\usepackage[small]{caption}
\usepackage{graphicx}
\usepackage{amsmath}
\usepackage{booktabs}
\usepackage{algorithm}
\usepackage{algorithmic}
\usepackage{subfig}
\usepackage{float}

\usepackage{mathbbol}

\title{Towards Quantifying Intrinsic Generalization of Deep ReLU Networks}

\author{
Shaeke Salman$^1$\and Canlin Zhang$^2$\and Xiuwen Liu$^1$\and Washington Mio$^2$\\
\affiliations
$^1$Department of Computer Science, Florida State University, FL 32306, USA\\
$^2$Department of Mathematics, Florida State University, FL 32306, USA\\
\emails
\{salman, liux\}@cs.fsu.edu,
czhang@math.fsu.edu, wmio@fsu.edu
}

\begin{document}

\maketitle

\begin{abstract}
Understanding the underlying mechanisms that enable the empirical successes of deep neural networks is essential for further improving their performance
and explaining such networks. 
Towards this goal, a specific question is how to explain the ``surprising" behavior of the same over-parametrized deep neural networks that can generalize well on real datasets and at the same time ``memorize" training samples when the labels are randomized. In this paper, we demonstrate that deep ReLU networks generalize from training samples to new points via piece-wise linear interpolation. We provide a quantified analysis on the generalization ability of a deep ReLU network: Given a fixed point $\mathbf{x}$ and a fixed direction in the input space $\mathcal{S}$, there is always a segment such that any point on the segment will be classified the same as the fixed point $\mathbf{x}$. We call this segment the $generalization \ interval$. We show that the generalization intervals of a ReLU network behave similarly along pairwise directions between samples of the same label in both real and random cases on the MNIST and CIFAR-10 datasets. This result suggests that the same interpolation mechanism is used in both cases. Additionally, for datasets using real labels, such networks provide a good approximation of the underlying manifold in the data, where the changes are much smaller along tangent directions than along normal directions. 
On the other hand, however, for datasets with random labels, generalization intervals along mid-lines of triangles with the same label are much smaller than those on the datasets with real labels, suggesting different behaviors along other directions. 
When the dimension of the input space is higher, the training samples are more sparse, and the differences are smaller. Our systematic experiments demonstrate for the first time that such deep neural networks generalize through the same interpolation and explain the differences between their performance on datasets with real and random labels.
\end{abstract}

\section{Introduction}

In recent years, deep neural networks have improved the state of the art performance substantially in computer vision~\citep{AlexNet2012Imagenet,DeepResidual2016He,Salman2019Sparsity}, machine translation~\citep{SeqToSeq2014Sut}, speech recognition~\citep{SpeechReco2013Alex}, healthcare~\citep{Miotto2017HealthDL,Salman2019Consensus-based} and game playing~\citep{AlphaGo2017Silver2017} among other applications. However, the underlying mechanisms that enable them to perform well are still not well understood. Even though they typically have more parameters than the training samples and exhibit very large capacities, they generalize well on real datasets trained via stochastic gradient descent or its variants.  In an insightful paper,~\citeauthor{Zhang2016UnderstandingDL}~[\citeyear{Zhang2016UnderstandingDL}] have identified a number of intriguing phenomena of such networks. In particular, they demonstrate that over-parametrized neural networks can achieve $100\%$ accuracy trained on datasets with the original labels and generalize well. At the same time, the exact same neural network architectures can also achieve $100\%$ accuracy on the datasets with random labels, and therefore ``memorize" the training samples. 
Clearly, this is not consistent with statistical learning theory~\citep{Stat1998Vapnik} and bias-variance trade-off~\citep{Geman1992Bias}, where models should match the (unknown) capacity of the underlying processes in order to generalize well.
Understanding and explaining this typical behavior of deep neural networks has attracted a lot of attention recently with the hope of revealing the underlying mechanisms of how deep neural networks generalize.

Fundamentally, while training, deep neural networks iteratively minimize a loss function defined as the sum of the loss on the training samples. The parameters in the trained network depend on the initial parameter values, the optimization process, and training data.
As 100\% accuracy on the training samples can be achieved even with random labels, finding 
good solutions for such networks that minimize the loss is therefore not a key issue.
While regularization techniques can affect the parameters of trained networks,  \citeauthor{Zhang2016UnderstandingDL} [\citeyear{Zhang2016UnderstandingDL}] have demonstrated that their effects are typically small, suggesting
that they are not a key component. 
Therefore, the generalization performance of a trained over-parametrized network should depend on the 
training data and network architecture.
In this paper, we focus on deep ReLU networks.  
We show that such networks generalize consistently and reliably by interpolating among the training points. Using generalization intervals defined as the range of the data that have the same classification along a direction, we discover that pairwise generalization intervals on datasets with real and random labels are almost identical for high dimensional inputs (e.g., MNIST and CIFAR-10 samples). Furthermore, we show that
pairwise interpolations approximate the underlying manifold in the data well, enabling the networks
to generalize well. We show that the properties are remarkably consistent among networks with different
architectures and on different datasets. The properties enable us to characterize the generalization performance
of neural networks based on their behaviors on the training sets only, which we call intrinsic generalization.
This notion of generalization is very different from the typical definition of the performance gap on the training set and test set.
While intrinsic generalization of a network on a training set can be quantified through generalization intervals,
the gap-based generalization performance can not be studied without having a validation set or test set.
Furthermore, the gap-based definition is extrinsic as it can vary when a different validation set is used.
In other words, for the first time, we demonstrate 
the underlying mechanisms that enable over-parametrized networks to generalize
well when all the training samples are classified correctly.
The systematic results demonstrate the effectiveness of the proposed method and therefore validate the proposed solution. 


The rest of the paper is organized as follows. In the next section, we review recent works that are closely related to our study. After that, we present the theoretical foundation of the generalization mechanism via interpolation for deep ReLU networks. We introduce a novel notation called generalization interval (GI) to quantify the generalization of such networks. Then, we illustrate our proposed ideas on intrinsic generalization behavior of deep ReLU networks with systematic experiments on representative datasets such as MNIST and CIFAR-10 along with a two-dimensional synthetic dataset.  Finally, we discuss correlations between generalization intervals on training sets and validation accuracy and
whether there exists a mechanism in deep neural networks that supports deep memorization. We conclude the paper with a brief summary and plan for future work.

\section{Related Work}
The ability to perform well on new data is one of the hallmarks of any useful machine learning system,
including deep neural networks.
Traditionally, the statistical learning theory~\citep{Stat1998Vapnik} 
ties a model's ability to generalize well to whether its capacity matches the underlying (unknown) complexity of the problem. 
In contrast, in recent years, deep neural networks have empirically improved the performance of many tasks
even with more parameters than the number of training samples. 
 Understanding the generalization ability and mechanism of
 deep neural networks have become central to get their full practical benefit in real applications~\citep{GenError2018Jakubovitz}.
While important, it is difficult to study the generalization performance of a deep neural network as it is defined as the performance gap on a training set and test one. 

\citeauthor{Zhang2016UnderstandingDL}~[\citeyear{Zhang2016UnderstandingDL}] have demonstrated concretely that the same deep neural networks can generalize and at the same time can ``memorize" training set with random labels,
which makes it clear that the capacity of deep neural networks~\citep{AcloserLook2017Krueger} can not explain 
their ability to generalize.
Consequentially, a central research question is to discover the properties of deep networks, the data, or the interplay that allow the network to generalize well and at the same time ``memorize".

Mechanically, training deep neural networks is based on minimizing a loss iteratively using gradient descent from an initial, often randomized, solution. 
As deep neural networks implement continuous functions, 
similar inputs will produce similar outputs. 
If we use the literal definition of memorization, i.e., the learning process associates a particular input with a particular label~\citep{Collins2018DetectingMemo}, deep neural networks are not capable of memorization only. 
Therefore, deep neural networks must ``generalize" in that after training on a training set, they will
also produce answers for other inputs.
For example, for classification, a trained (deep neural network) model divides the input space into decision regions, and the inputs in the same region produce the same output. In other words, all the inputs are associated with 
the same output. 
As a result, for deep neural networks, the question becomes how they generalize, rather than whether they generalize. 

It appears that the discussions about memorization and generalization of deep neural networks are often
due to different definitions of memorization.
For example, in~\citep{AcloserLook2017Krueger,LearningAndMem18Chatterjee}, memorization is simply defined to be a point
when a network reaches $100\%$ accuracy on the training set. 
Note that for the network to generalize, it should perform well on the training set, i.e., it should memorize at the same time. This could not help elucidate the underlying mechanism that enables
a deep neural network to generalize.

In this paper, to quantify the intrinsic ability of a deep neural network to generalize, we introduce generalization intervals to measure how far a trained deep network can generalize. 
Through systematic experiments, we discover that pairwise generalization intervals are remarkably similar on datasets with real and random labels, confirming that generalization of deep neural networks is independent of the labeling of data. Furthermore, we show that on real datasets, pairwise interpolations provide a way to approximate the underlying manifolds that allow such models to generalize well.

\section{Theoretical Foundation for Deep ReLU Networks}

\subsection{Notations and ReLU Networks}

In this paper, we use the following commonly used notations. We focus on feed-forward deep neural networks having Rectified Linear Unit (ReLU) activations for classification which is defined by $t \mapsto max(0,t)$. The neural network functions can be defined as $f : \mathbb{R}^{d_x} \to \mathbb{R}^{d_y}$, where $d_x$ and $d_y$ are the input and output dimension respectively. We consider $x \in \mathbb{R}^{d_x}$ is an input vector and $y \in \mathbb{R}^{d_y}$ is a target vector. We define a set of n training samples as $\left\{(x_i,y_i)\right\}^n_{i=1}$. The functions $f$ are parameterized by $\theta$, which is a vector that includes all the parameters (i.e., weights and biases). Furthermore, we would use the function interchangeably to denote the pre-activation output of the last layer (i.e., output from the penultimate layer). 

By restricting the activation function to be ReLU, a neural network provides a piecewise approximation of complex decision regions~\citep{OnTheNumberofLinear14Montufar} by minimizing the empirical risk,
\begin{equation} \label{eqn0}
\min_{\theta} \frac{1}{n}\sum_{i=1}^{n}\ell(f(x_i;\theta),y_i),
\end{equation}
where $\ell: \mathbb{R}^{d_x} \times \mathbb{R}^{d_y} \to \mathbb{R}$ denotes some loss function such as $\ell_2$ loss, hinge loss, cross entropy loss etc. The gradient based methods update the weights by 
\begin{equation} \label{eqngd}
W = W - \eta \sum_{i=1}^{n}\nabla_W\ell(f(x_i;\theta),y_i),
\end{equation}
where $\eta$ is known as the step size. To simplify our classification model, we have considered the cross entropy loss. Let $l$ denotes a particular layer, $L$ the last layer of the network, k the index of a neuron and 
c the number of classes.
Given that the last layer is a softmax layer, the activations for the neurons in the last layer would be defined by
\begin{equation}
    a_k^L = softmax(z_k^L) = \frac{e^{z_k^L}}{\sum_{j=1}^{c} e^{z_j^L}},
\end{equation}
where $z_k^L$ denotes weighted summation of activations from prior layer. The cross entropy loss is defined as follows. 
\begin{equation} \label{ce}
   e = -\sum_i y_i \log a_k^L
\end{equation}

Now, from the definition of the loss, we can derive the weight updates for the last layer as the following.
\begin{equation} \label{competitive}
   \frac{\partial e}{\partial W_{kj}^L} = -r_p^{L-1}\sum_i (y_i- \frac{e^{z_k^L}}{\sum_{j=1}^{c} e^{z_j^L}}),
\end{equation}
where $r_p^{L-1}$ denotes the ReLU activations of previous layer. 
Note that $y_i$ is 1 for the target class (which can be a real label or a random label assigned to the sample) and 0 for
all the other classes.
Equation \ref{competitive} shows the importance
of labels for gradient descent. The weights 
are getting closer to the representation of the neuron in the assigned
class in the previous layer while pushing away from
that of the neurons in the other classes,
representing competitive learning~\citep{Competitive98Rumelhart}.



\subsection{Generalization Intervals}

In order to study generalization behavior quantitatively, we introduce the Generalization Interval (GI), which is defined
as the generalization interval of a trained deep neural network at input point $x_0$ along a particular direction $d$ as the following.

\begin{equation} \label{eqngi}
\begin{aligned}
GI(f(x;\theta), x_0, d) ={} & \sup\{\epsilon \colon (f(x_0 + td);\theta) =\\ 
&f(x_0;\theta), 0\leq t \leq \epsilon\} + \sup\{\epsilon \colon (f(x_0 -\\ &td);\theta) = f(x_0;\theta), 0\leq t \leq \epsilon\}
\end{aligned}
\end{equation}

Intuitively,
the above equation defines the generalization interval at a given point as the range of the inputs that have stable classification along a particular direction.
When the samples are from a meaningful data distribution, i.e., when the samples are from an underlying manifold, the generalization intervals along the tangent directions of the manifold should be large. 
By connecting the samples with the same true labels in a pairwise manner,
one can approximate the tangent directions smoothly of the manifold. 
In contrast, the directions with other labels approximate the normal directions generally, resulting in poor generalizations (i.e., small generalization intervals).

Since there are many such pairwise paths, a question is whether multiple 
lines interfere with each other. 
In a high dimensional space (such as MNIST or CIFAR-10; see the Experiments Section), 
the probability of two lines intersecting with the other lines or getting very close to other lines is low. 
However, when the dimension gets smaller, the chance of having two lines interfering with each other will be higher. 
In addition, the interference affects the performance of a trained neural network only if the two
points are from different classes. 
On datasets with  random labels, one would expect higher chances of interference if closer points with different 
labels are much more common.
However, we argue that deep neural networks still do not simply memorize training points even 
when the labels are randomized. 
Please see the Experiments Section for numerical results.


\subsection{Generalization Along Tangent and Normal Directions}

\begin{figure*}[ht]
  \centering
  \vspace{-0.15in}
  \includegraphics[width=0.33\textwidth]{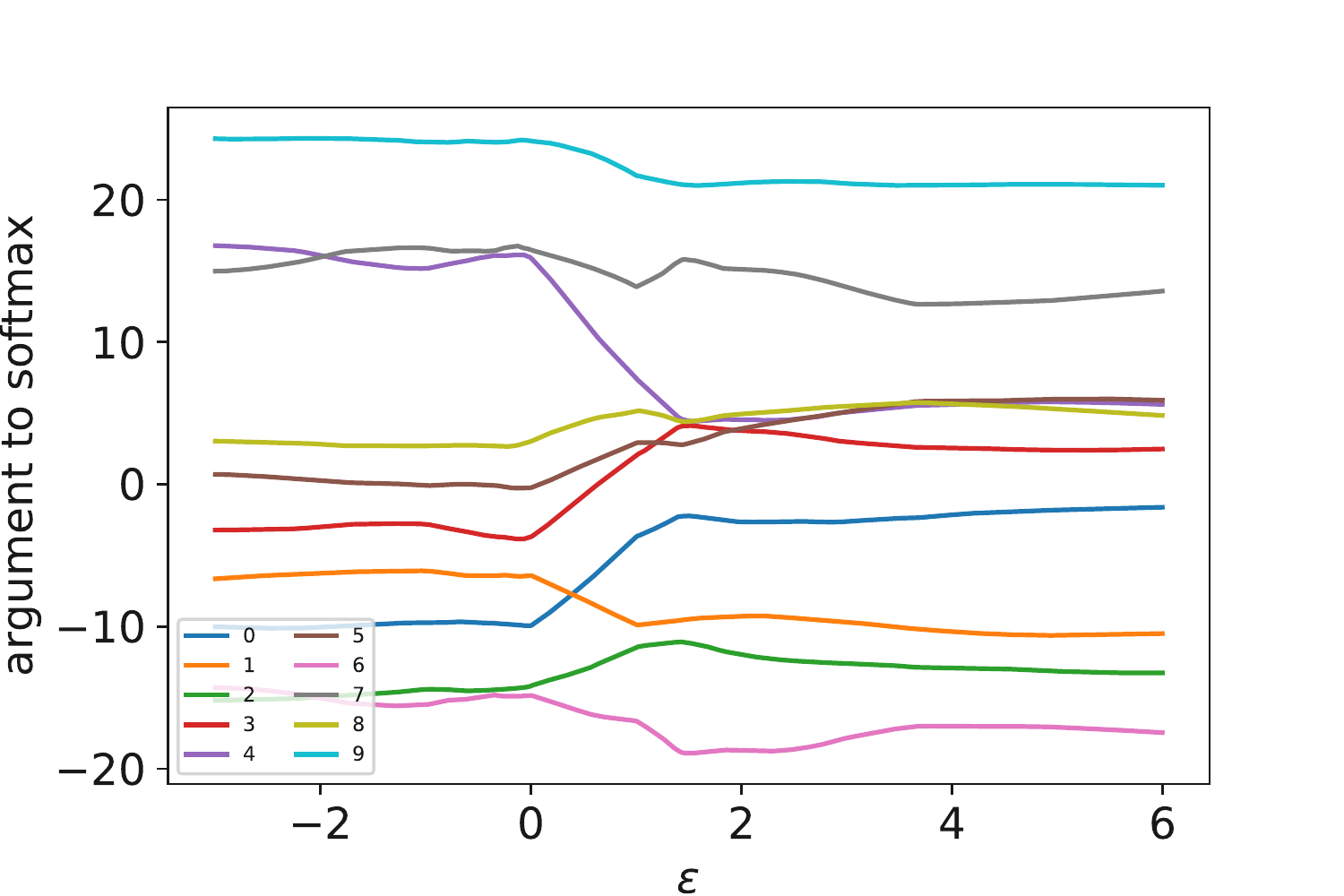}
  \includegraphics[width=0.33\textwidth]{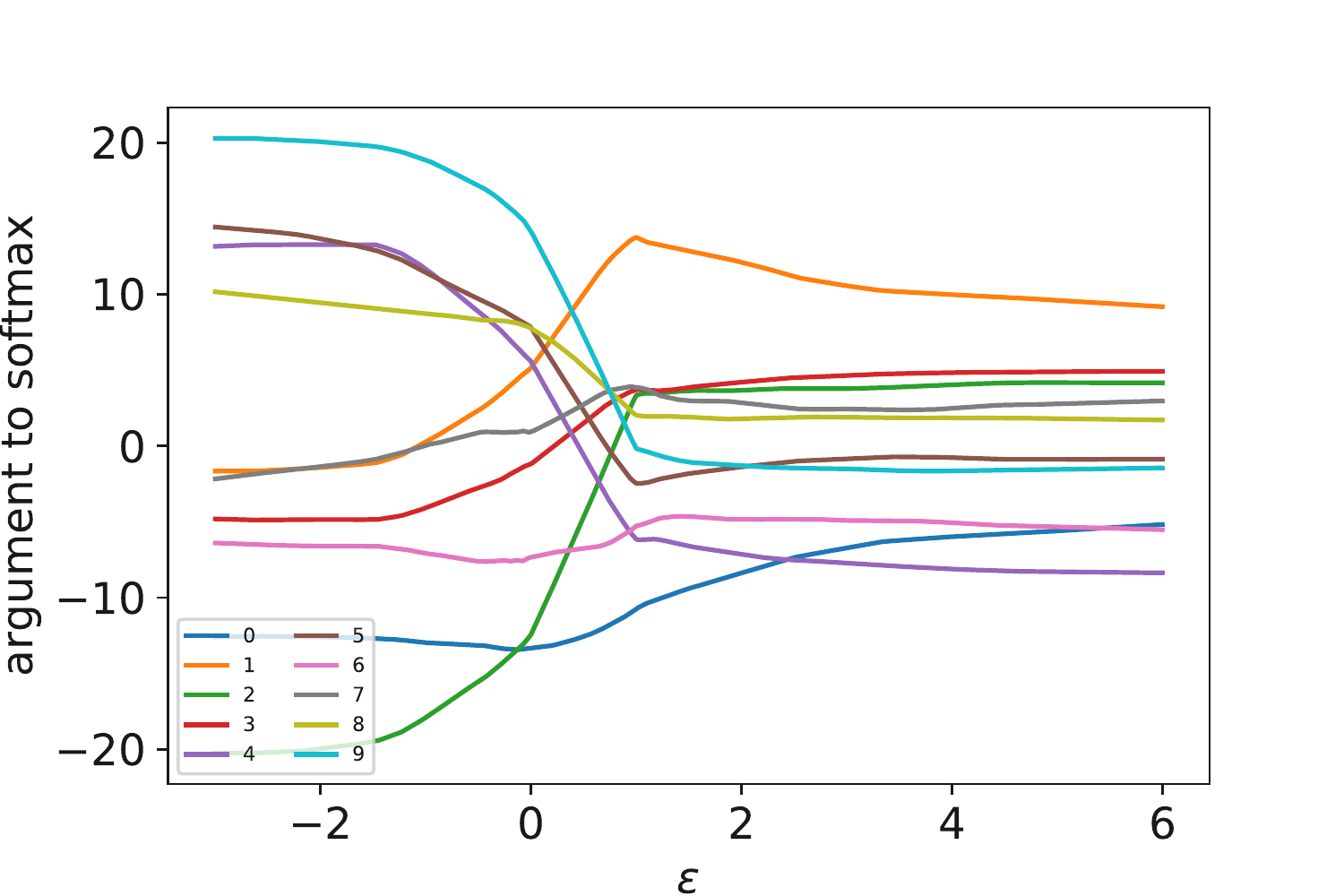}
  \includegraphics[width=0.33\textwidth]{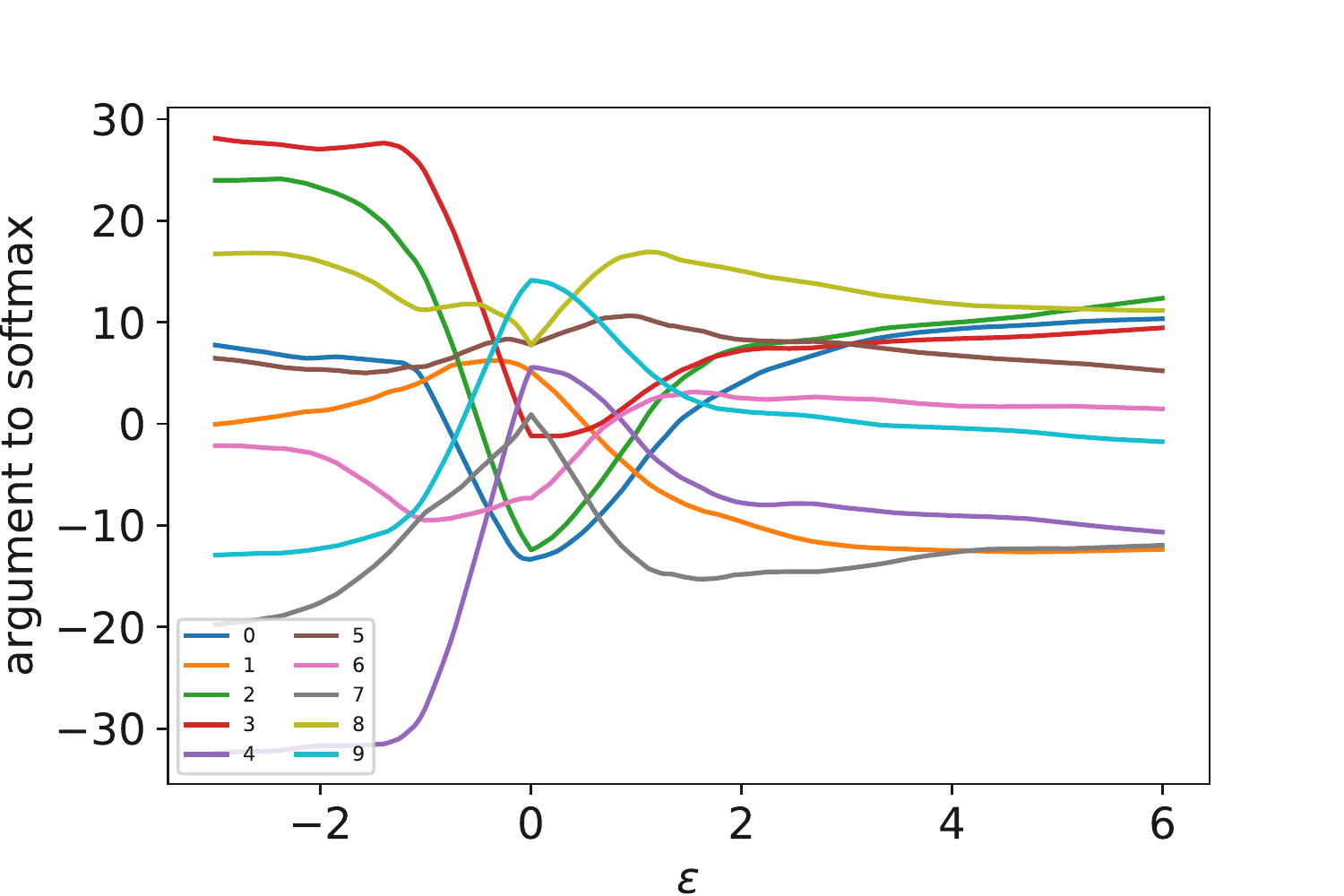}
  \vspace{-0.10in}
  \caption{(to be viewed in color) Outputs from the penultimate layer when network is trained on real labels, along a certain direction. (left) from one sample to another sample in the same class (class 9). (middle) from one sample (class 9) to another sample in a different class (class 1). (right) random direction. 
  }
  \label{fig:tangent_normal}
\end{figure*}

\begin{figure*}[ht]
  \centering
  \vspace{-0.15in}
  \includegraphics[width=0.33\textwidth]{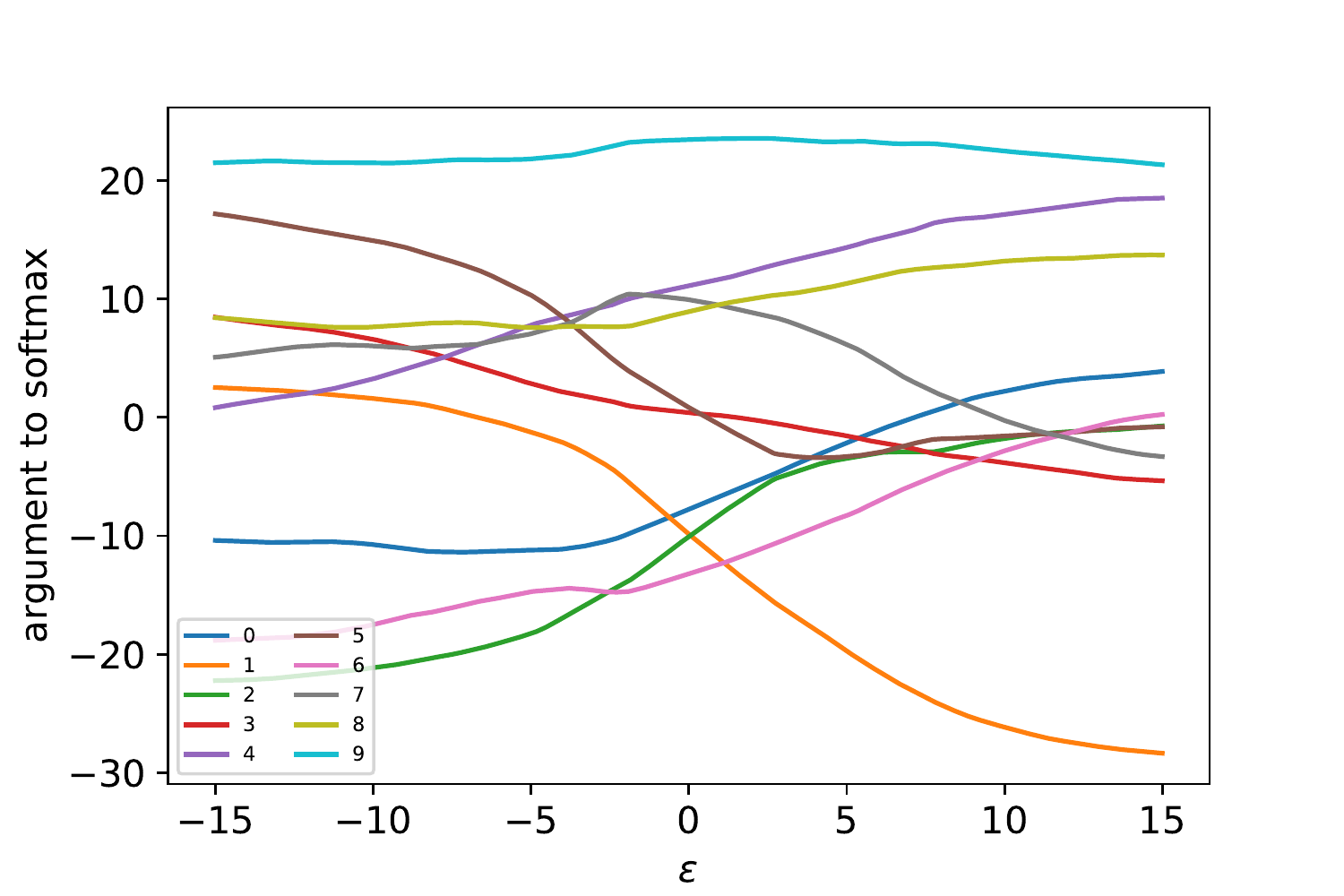}
  \includegraphics[width=0.33\textwidth]{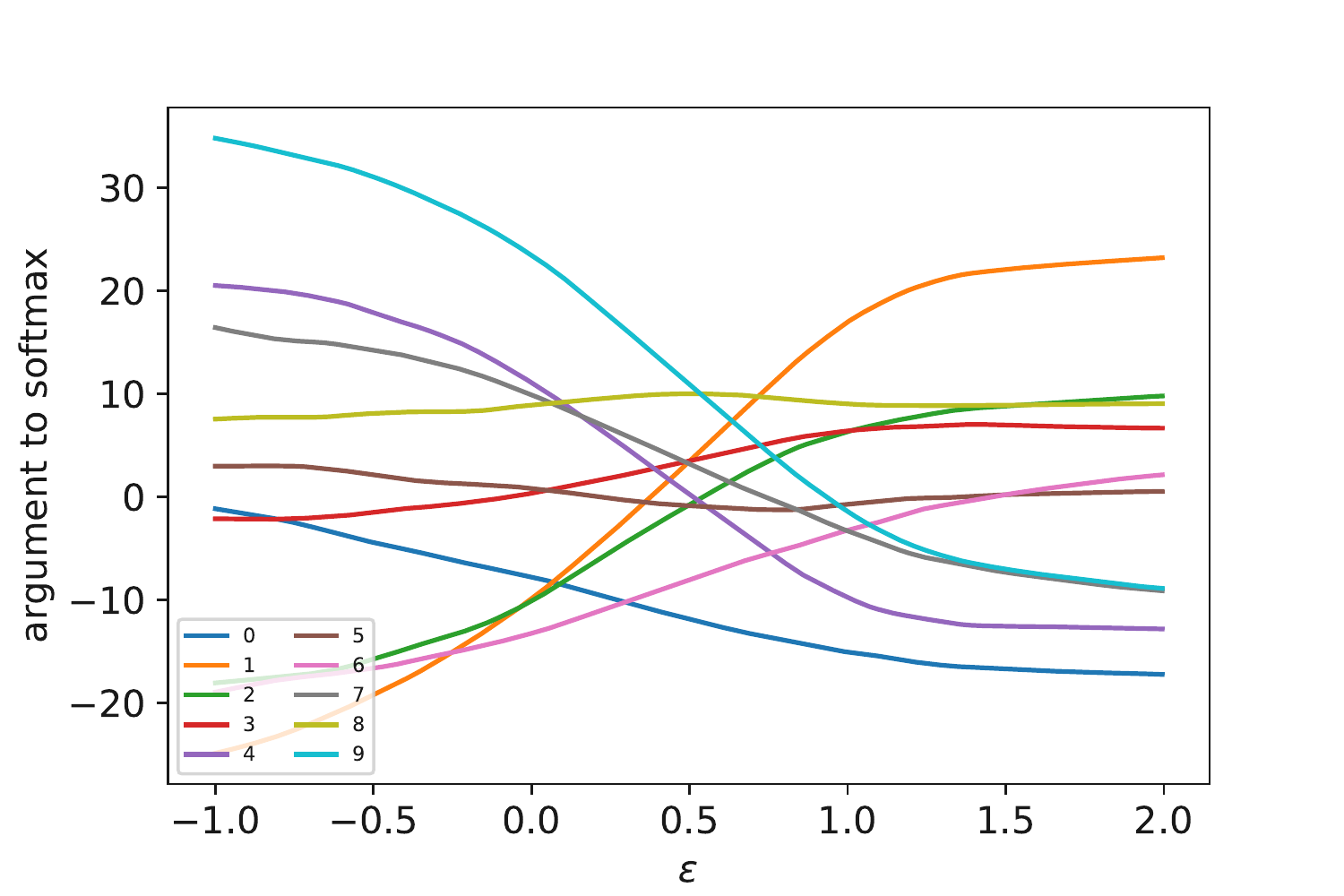}
  \includegraphics[width=0.33\textwidth]{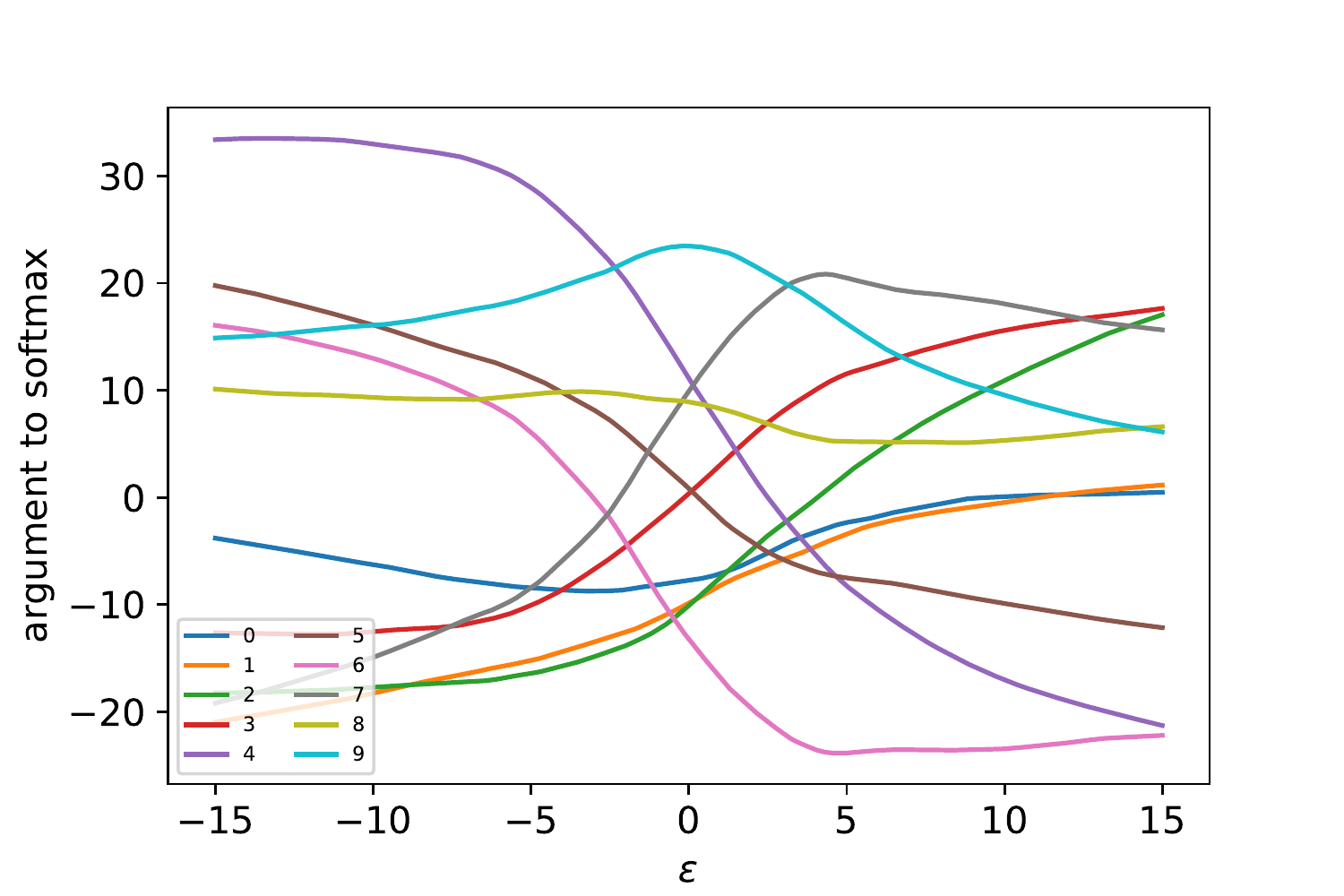}
  \vspace{-0.10in}
  \caption{(to be viewed in color) When the models are trained with real labels, outputs from the penultimate layer along the direction defined by class 9 samples. (left) first principal component. (middle) perpendicular (mean class 9 to mean class 1) to the first principal component. (right) principal component having a small eigenvalue (i.e., approximately a normal direction). 
  }
  \label{fig:tangent_normal_pca}
\end{figure*}

\begin{figure*}[ht]
  \centering
  \vspace{-0.15in}
  \includegraphics[width=0.33\textwidth]{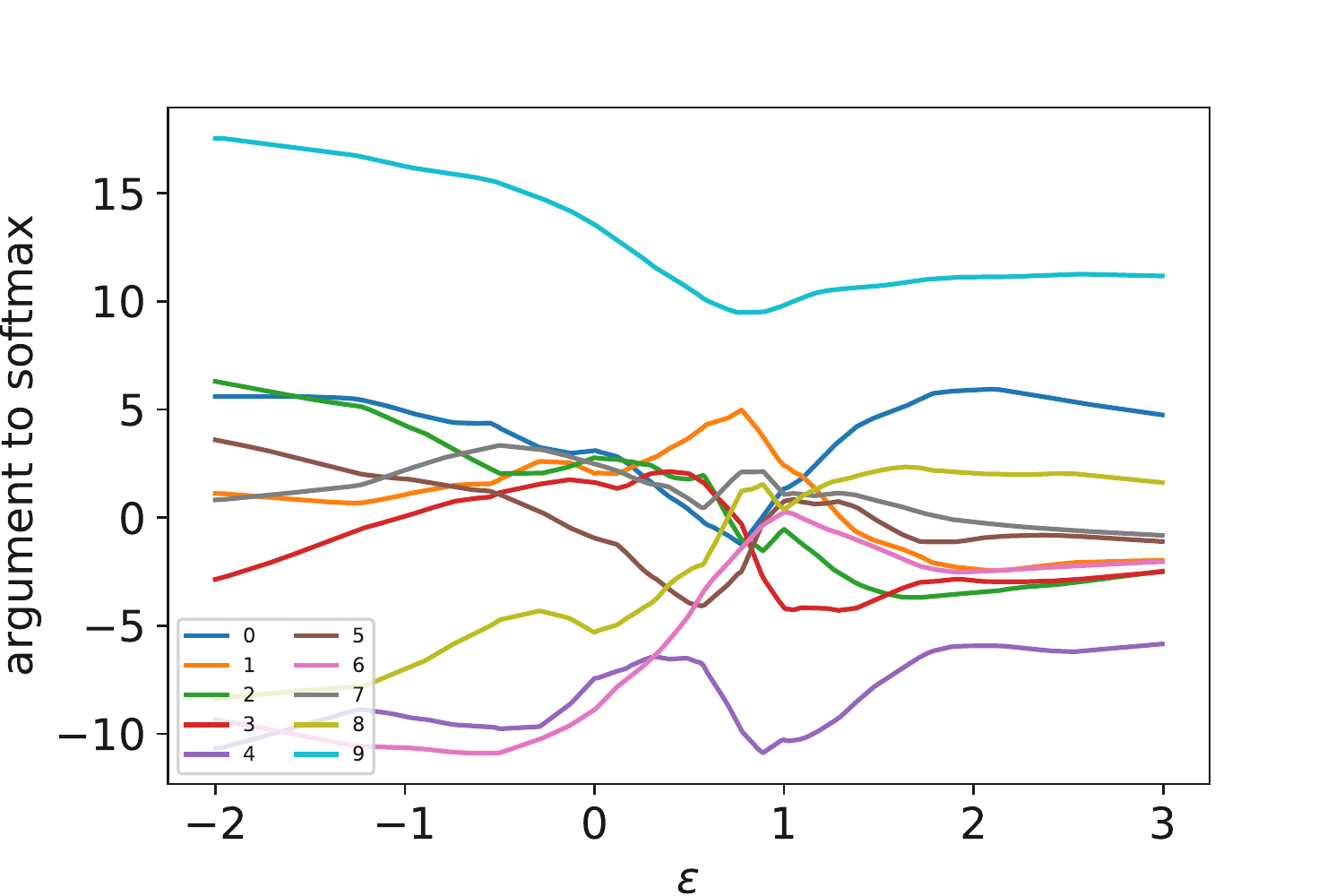}
  \includegraphics[width=0.33\textwidth]{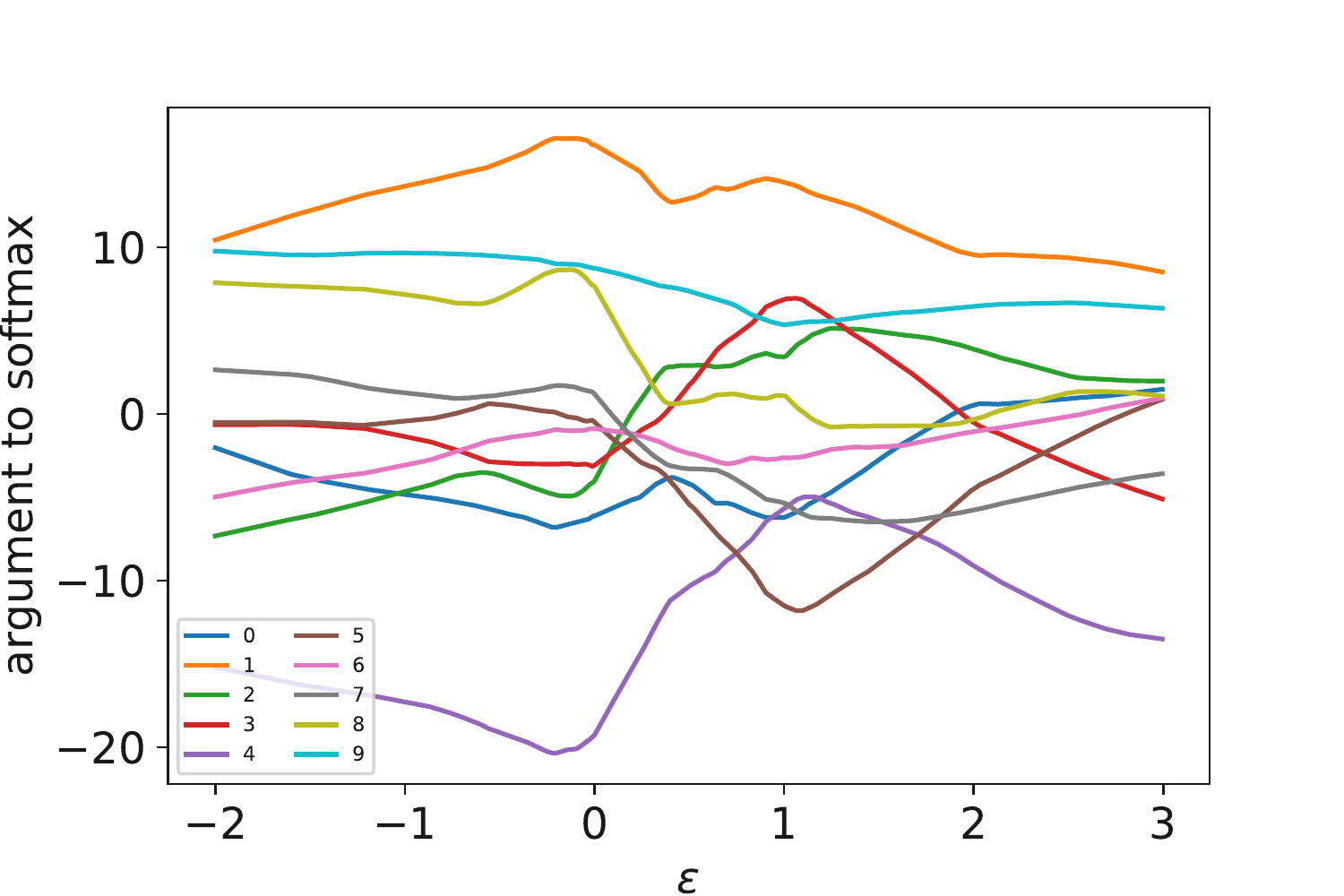}
  \includegraphics[width=0.33\textwidth]{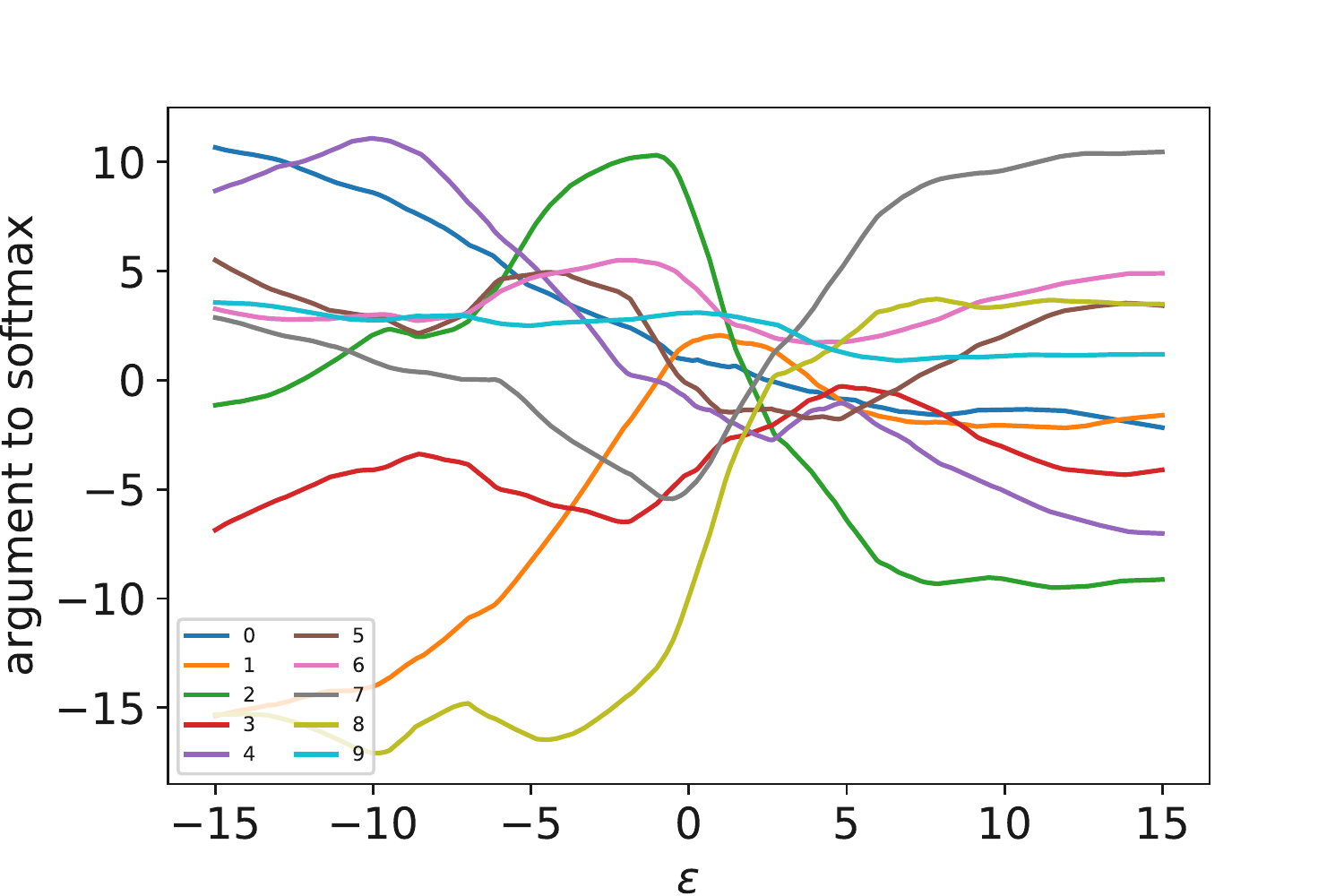}
  \vspace{-0.10in}
  \caption{(to be viewed in color) When the models are trained with random labels, outputs from the penultimate layer along the direction defined by (left) two samples in the random class 9. (middle) two samples in the random class 1. (right) first principal component of random class 9. 
  }
  \label{fig:rl_same_random_class_pca_1st}
\end{figure*}

\begin{figure*}[ht]
  \centering
  \vspace{-0.15in}
  \includegraphics[width=0.33\textwidth]{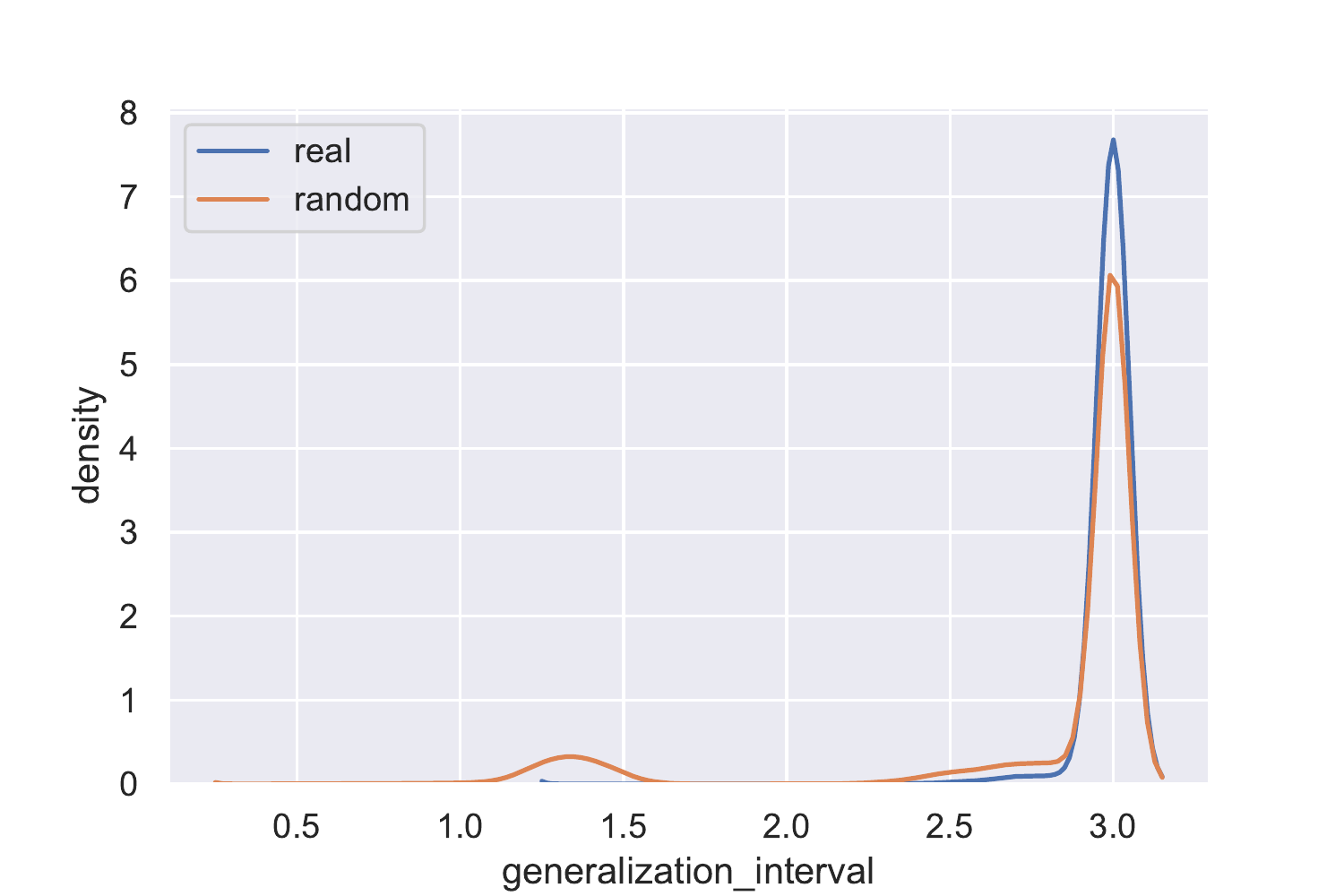}
  \includegraphics[width=0.33\textwidth]{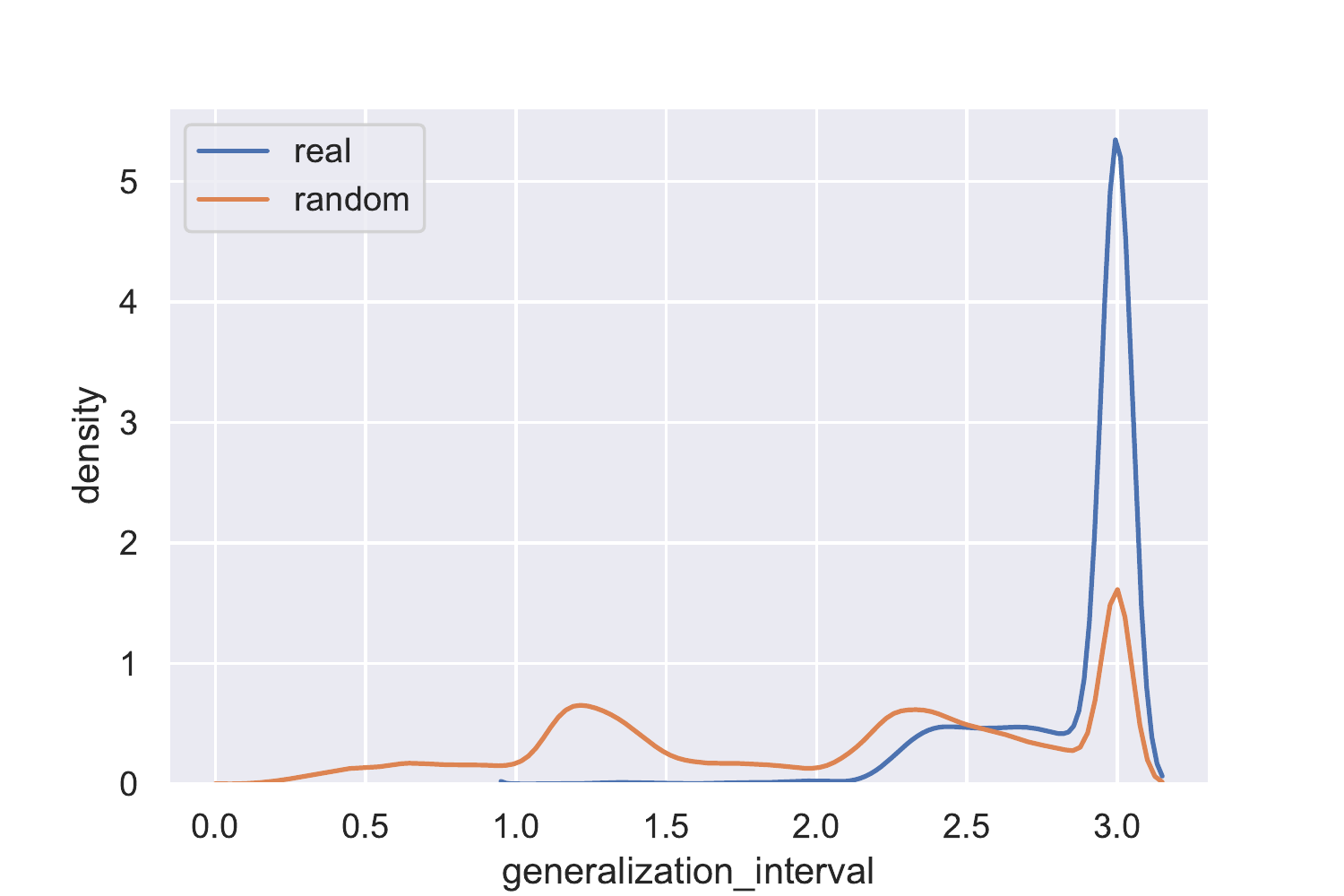}
  \includegraphics[width=0.33\textwidth]{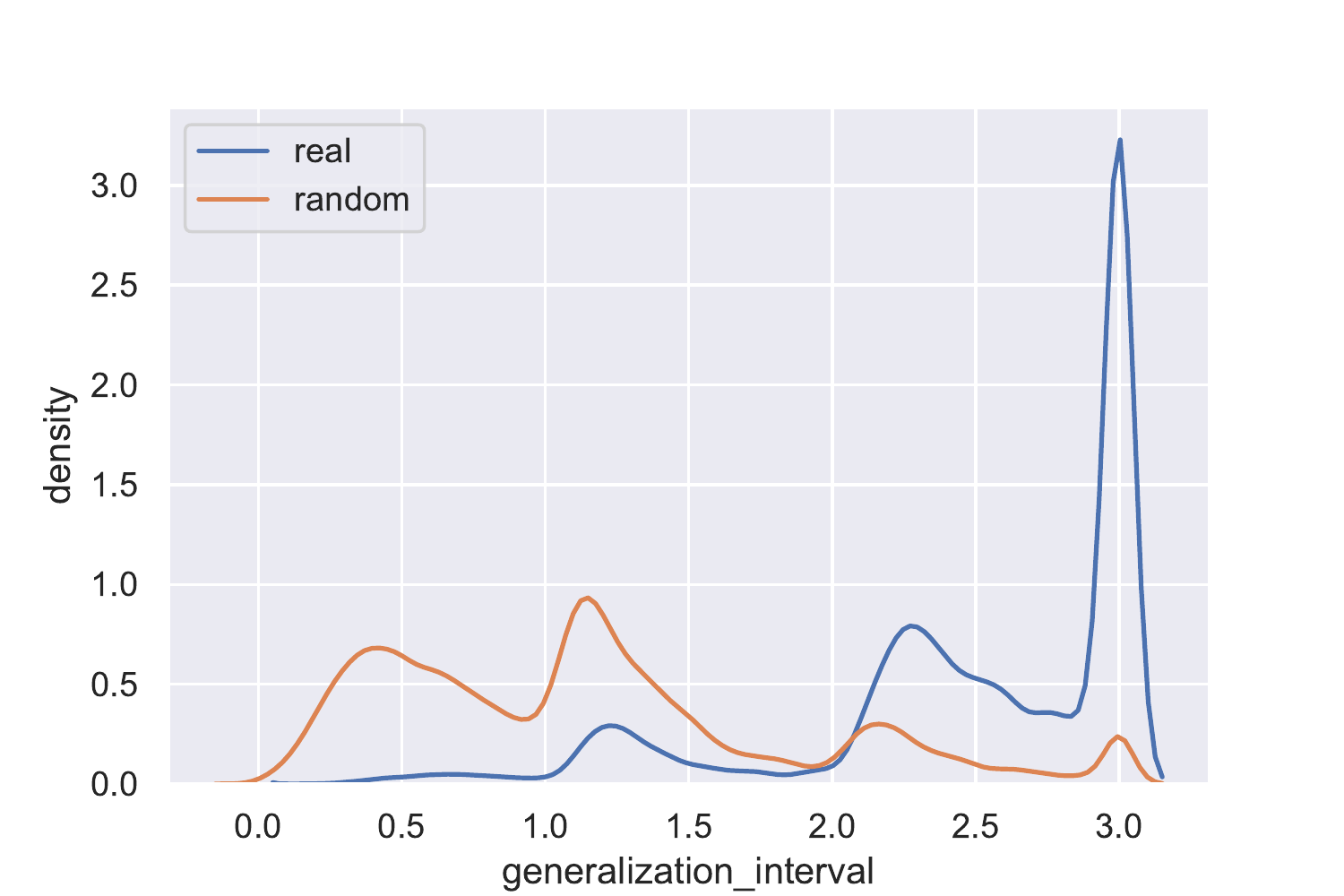}
  \vspace{-0.10in}
  \caption{Generalization intervals density plots for lines defined by the samples of MNIST having real and random labels of class 5. (left) dimension unchanged. (middle) down-sampled to 14$\times$14. (right) down-sampled to 7$\times$7.}
  \label{fig:mnist_lines_class_5}
\end{figure*}

\begin{figure*}[ht]
  \centering
  \vspace{-0.15in}
  \includegraphics[width=0.33\textwidth]{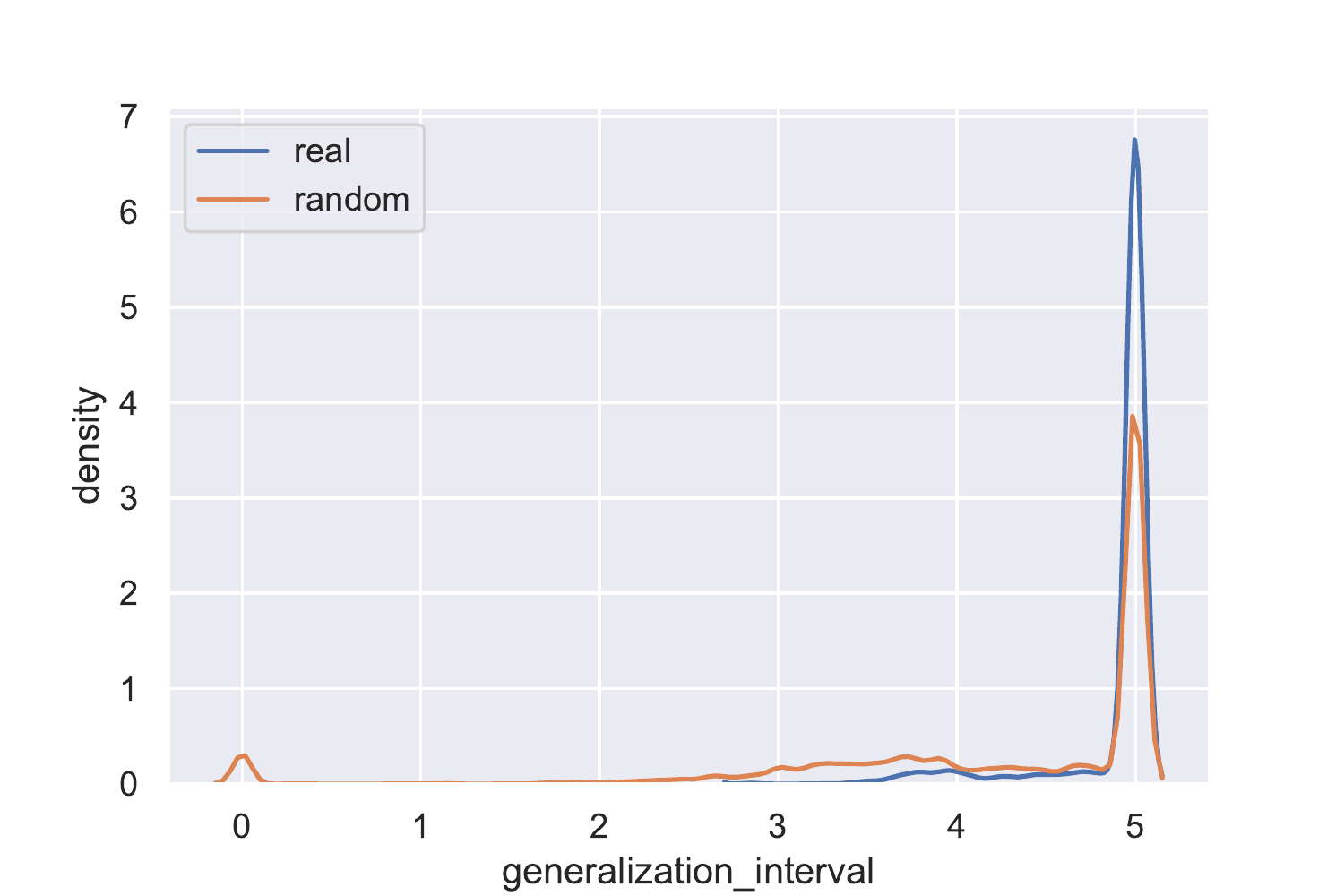}
  \includegraphics[width=0.33\textwidth]{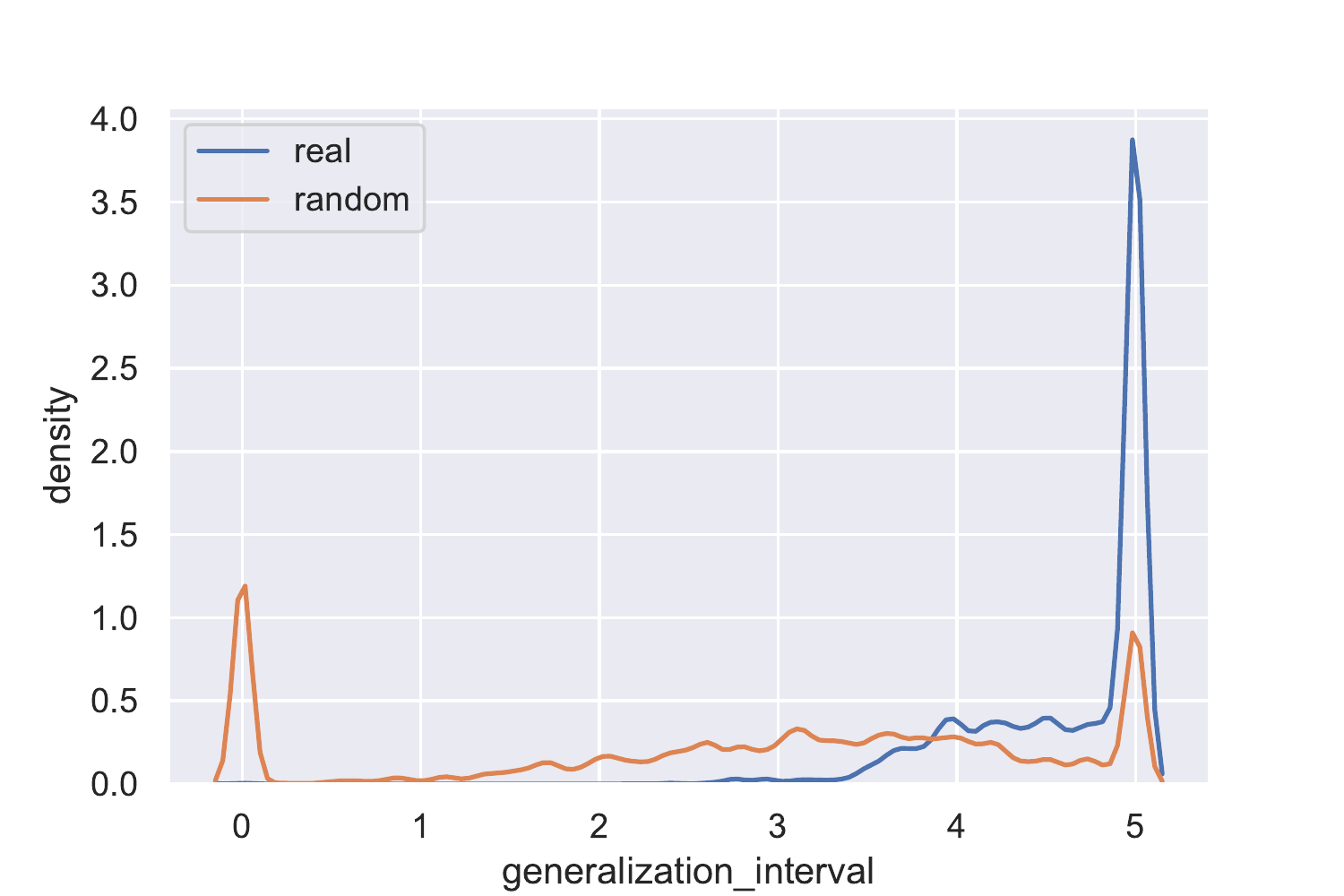}
  \includegraphics[width=0.33\textwidth]{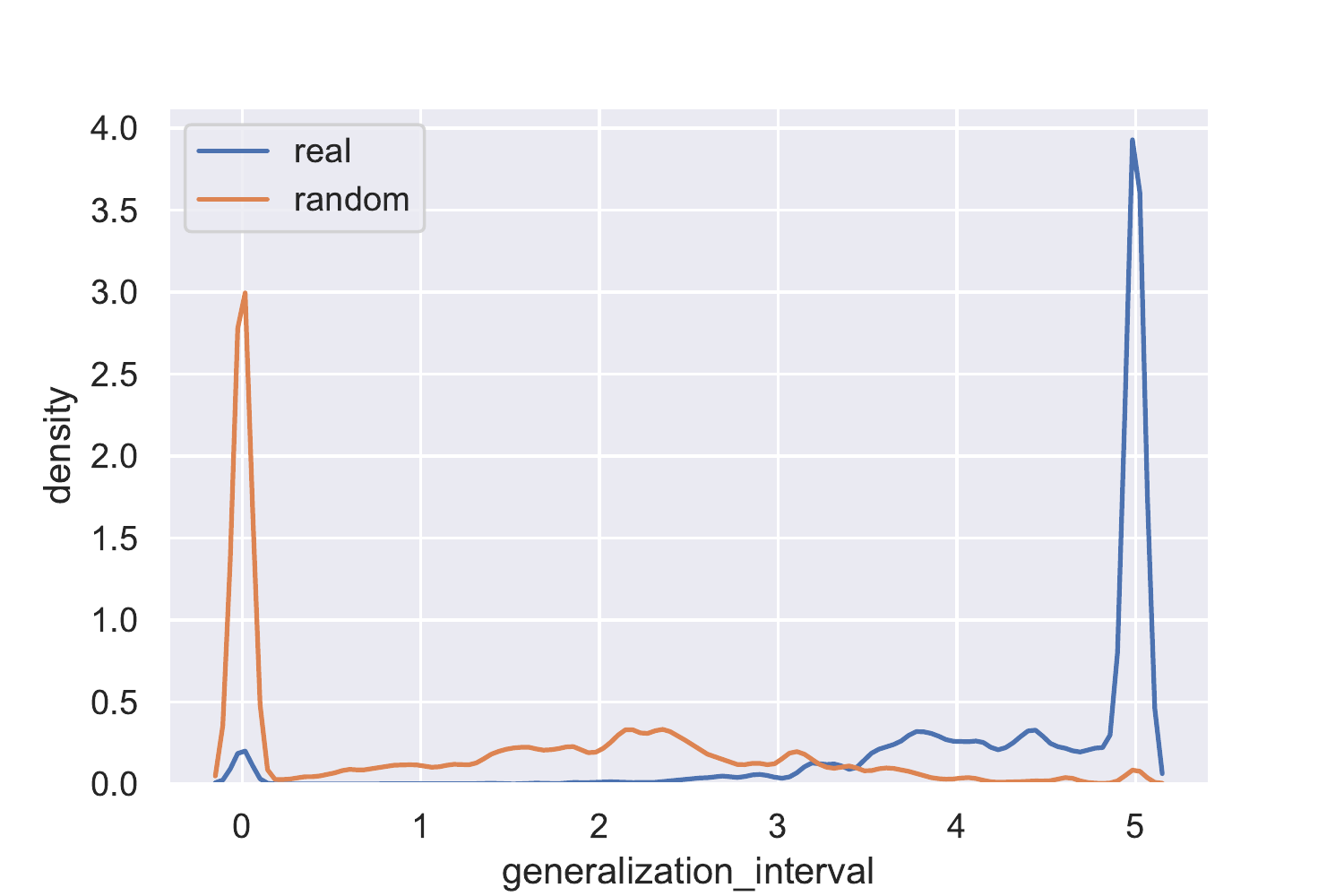}
  \vspace{-0.10in}
  \caption{Generalization intervals density plots for connecting midpoints of triangles defined by the samples of MNIST having real and random labels of class 5. (left) dimension unchanged. (middle) down-sampled to 14$\times$14. (right) down-sampled to 7$\times$7.}
  \label{fig:mnist_triangles_class_5}
\end{figure*}

\begin{figure*}[ht]
  \centering
  \vspace{-0.15in}
  \includegraphics[width=0.33\textwidth]{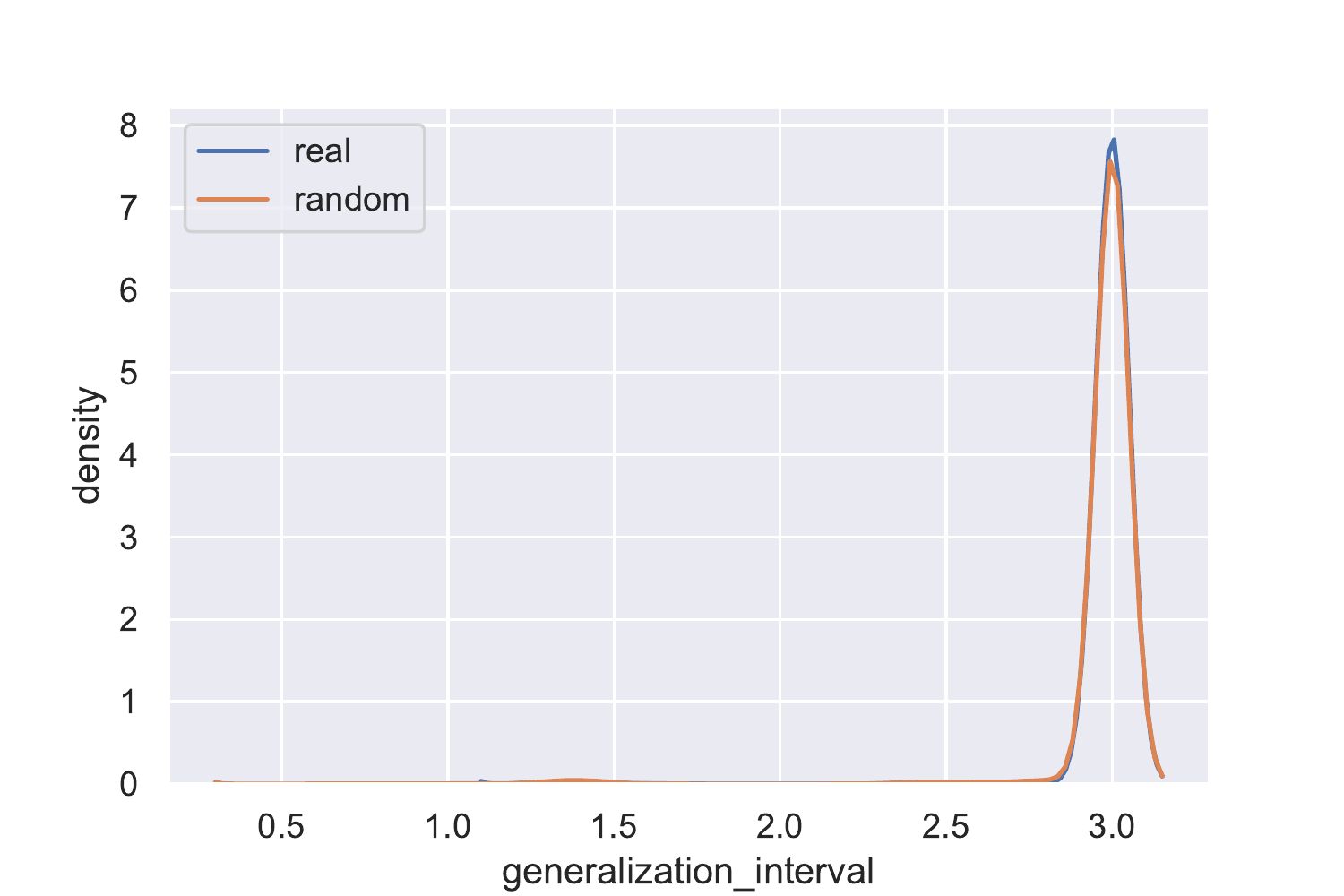}
  \includegraphics[width=0.33\textwidth]{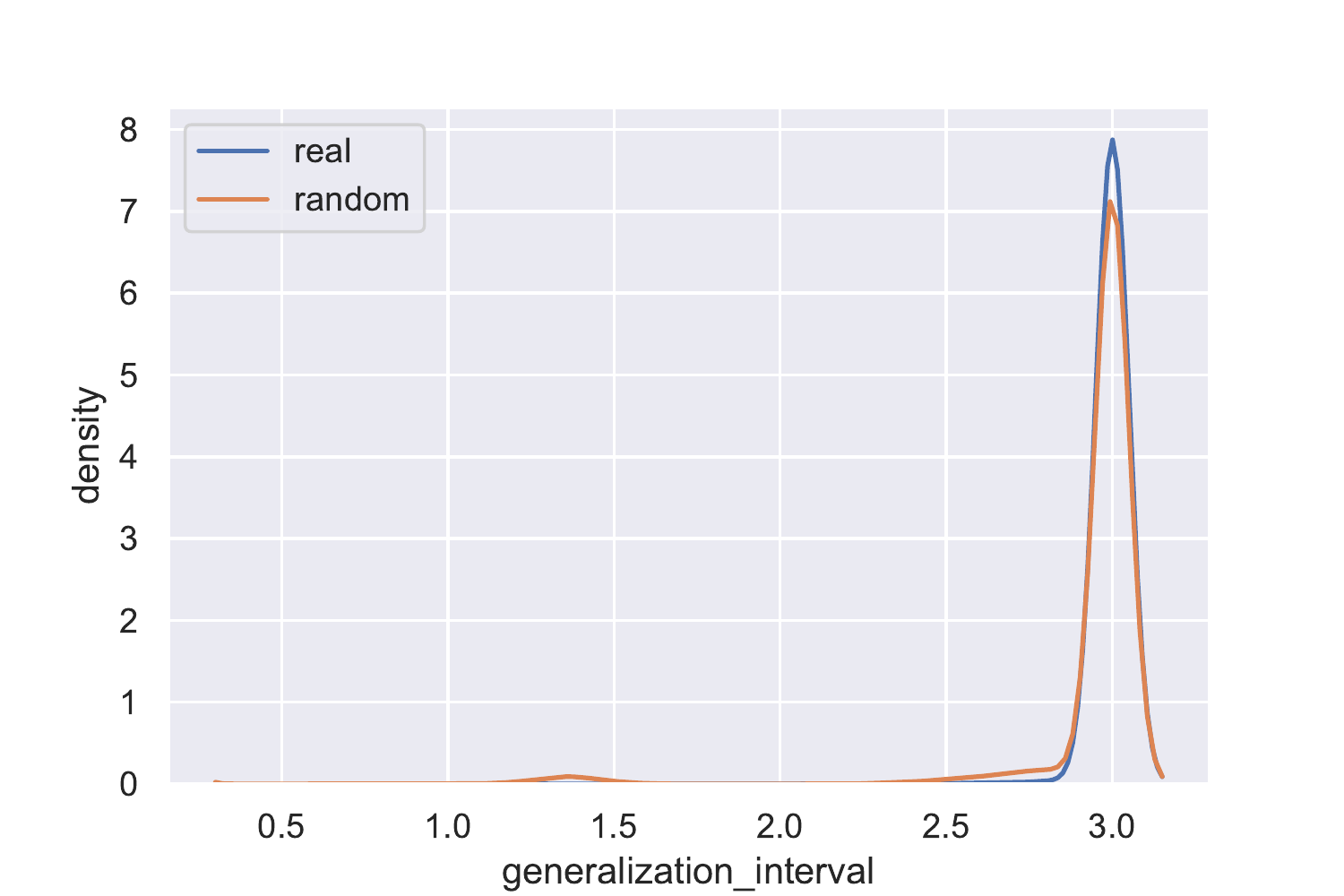}
  \includegraphics[width=0.33\textwidth]{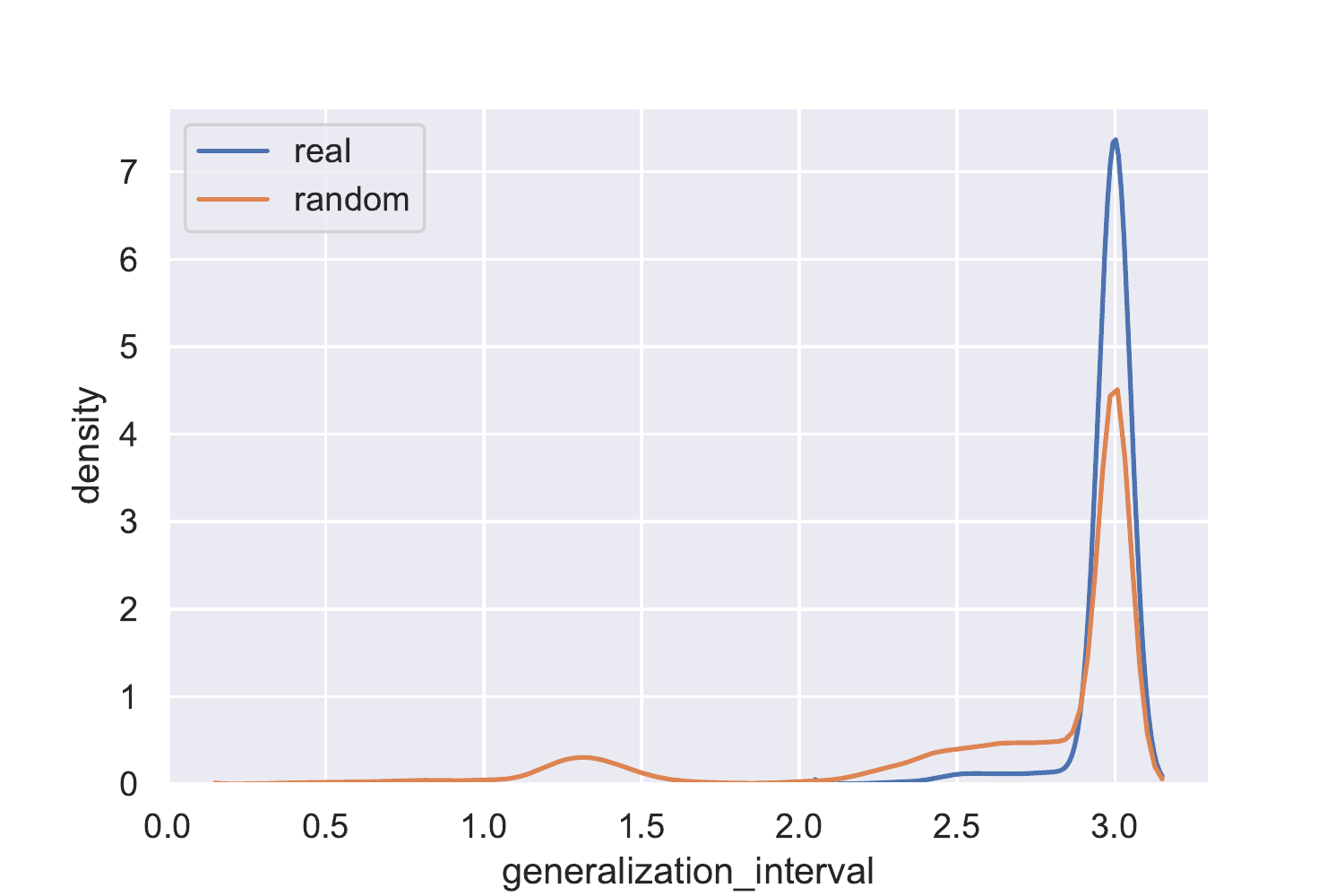}
  \vspace{-0.10in}
  \caption{Generalization intervals density plots for lines defined by the samples of MNIST having real and random labels of class 5 and for different architectures with (left) 1 hidden layer. (middle) 2 hidden layers. (right) 4 hidden layers.}
  \label{fig:mnist_lines_class_5_diff_layers}
\end{figure*}

\begin{figure}[ht]
    \centering
    \vspace{-0.15in}
    \includegraphics[width=7cm]{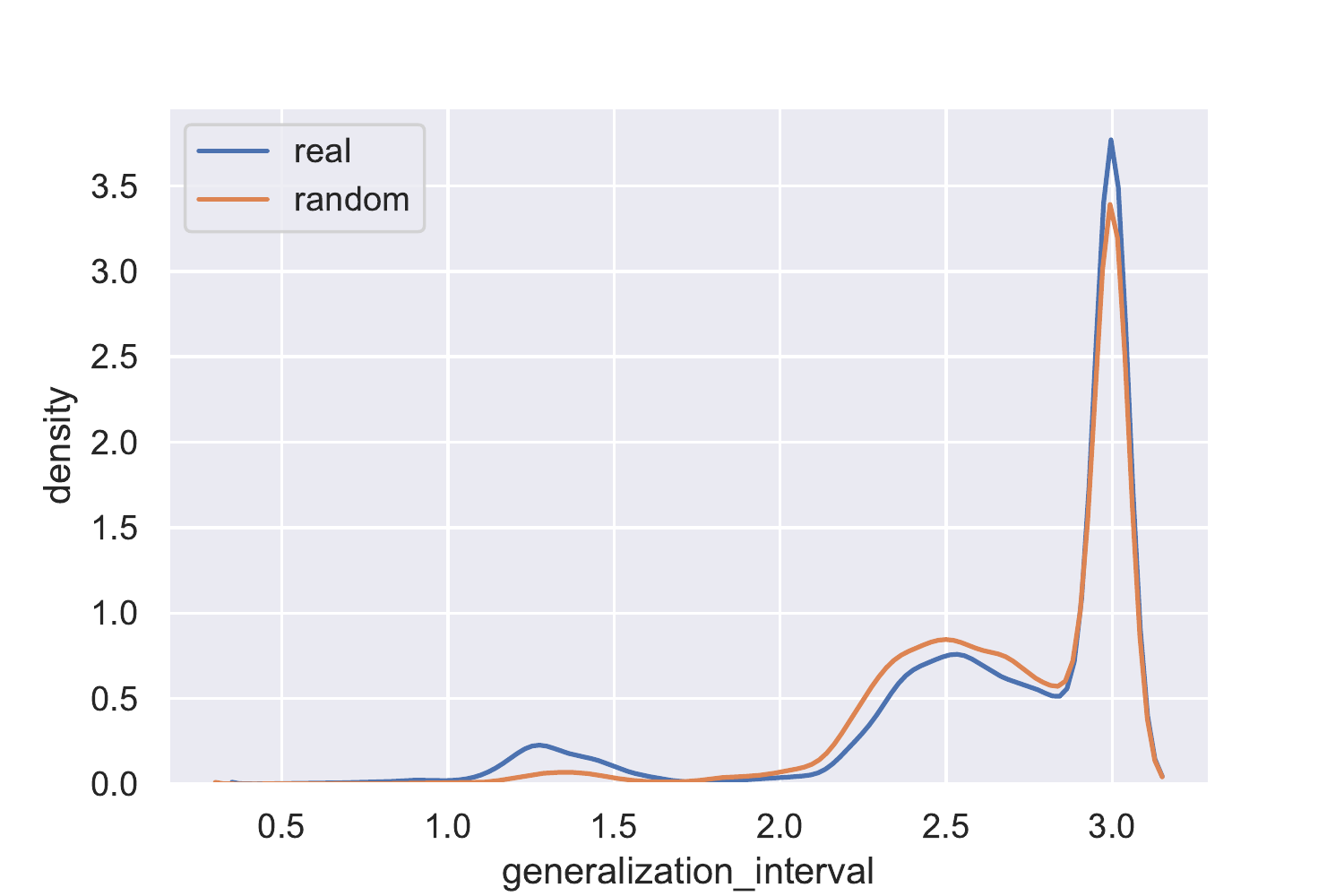}
    \vspace{-0.10in}
    \caption{Generalization intervals density plots for lines defined by the CIFAR-10 samples of having real and random labels of class 5.}
    \label{fig:gi_cifar10_al_rl}
\end{figure}

\begin{figure}[ht]
    \centering
    \vspace{-0.15in}
    \includegraphics[width=7cm]{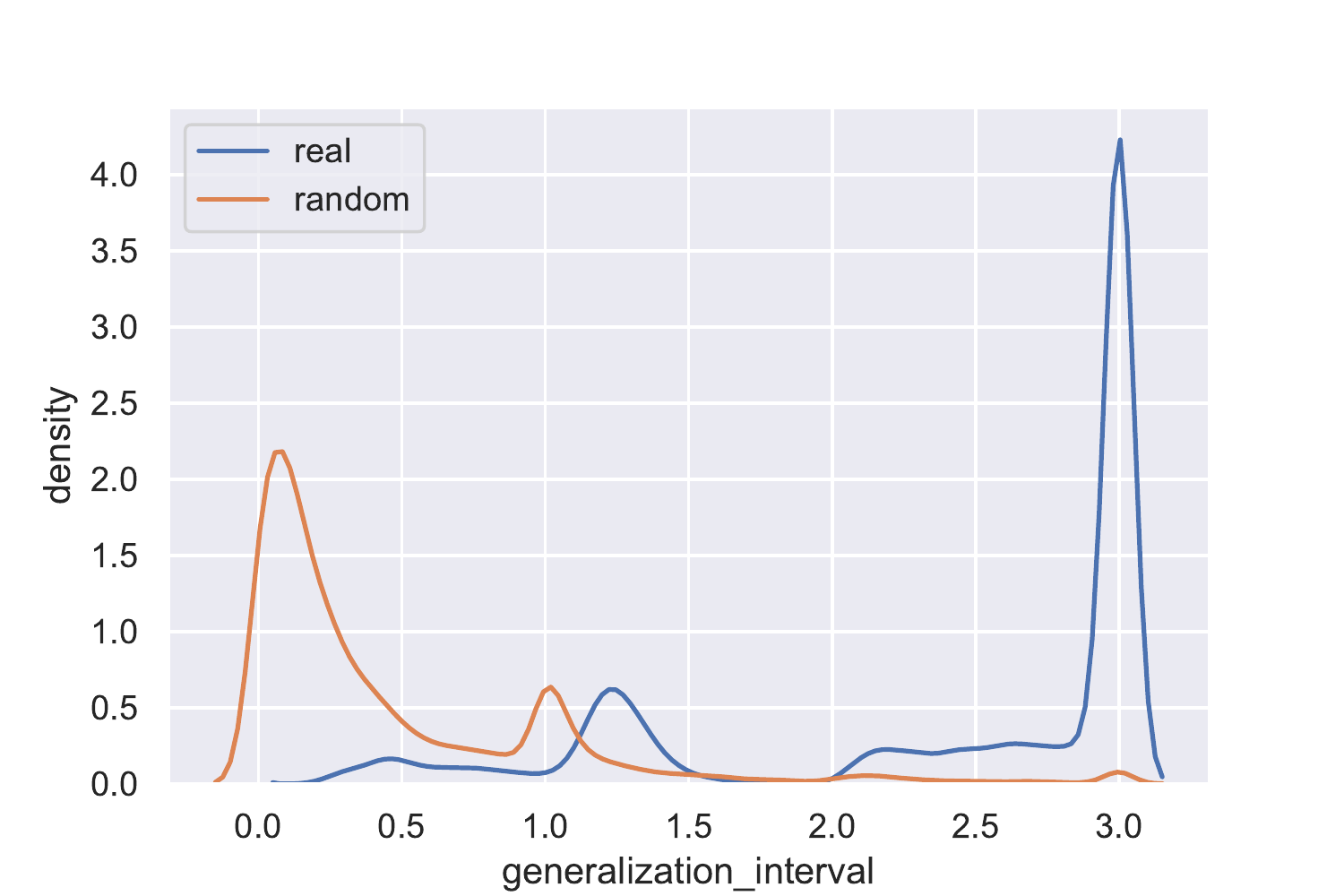}
    \vspace{-0.10in}
    \caption{Generalization intervals density plots for lines defined by the two-dimensional ``two moons" samples of having real and random labels of class 0 (red).}
    \label{fig:gi_2d_al_rl}
\end{figure}

\begin{figure}[ht]
  \centering
  \vspace{-0.15in}
  \includegraphics[width=0.450\textwidth]{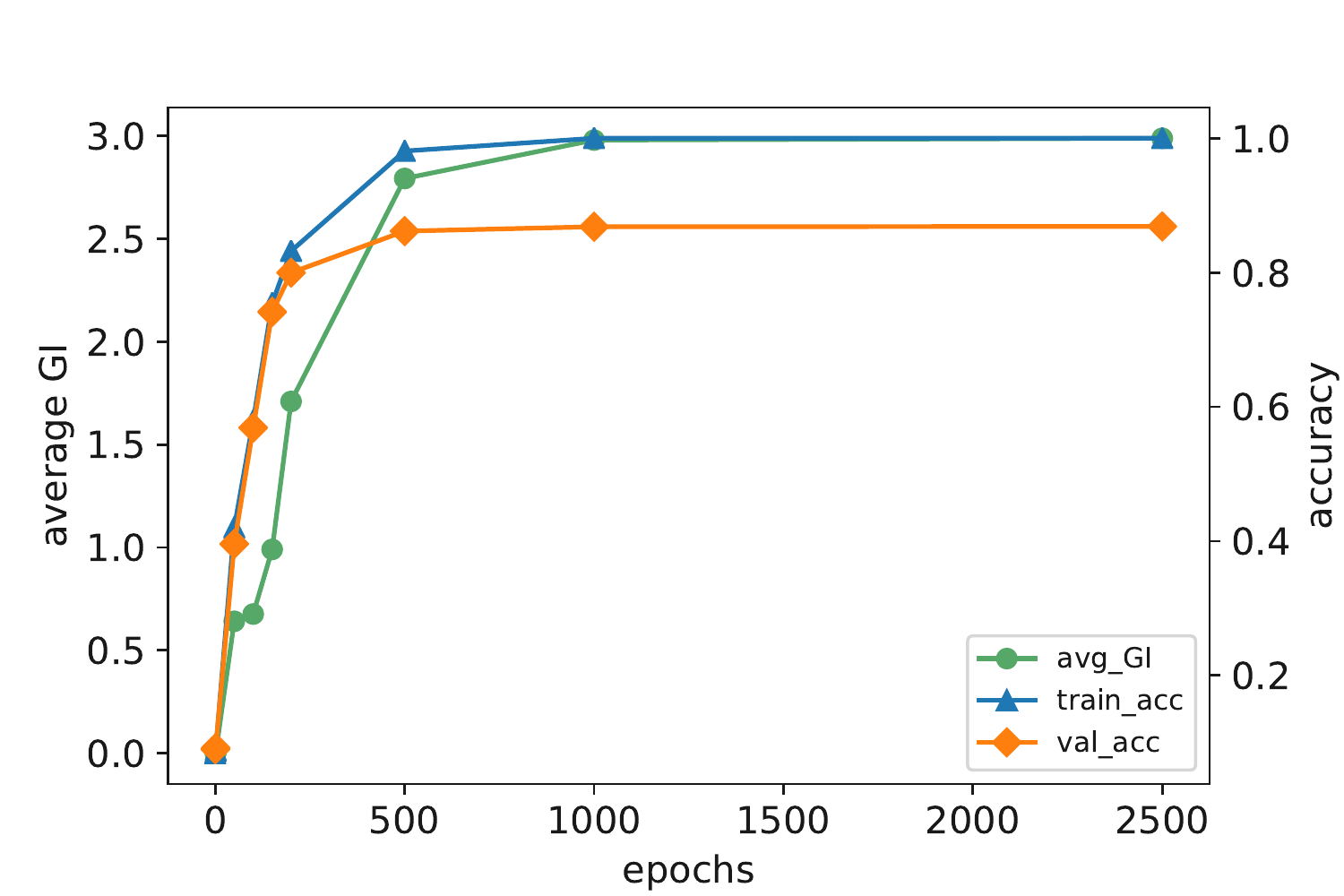}\label{fig:mnist_corr_2828}
  \vspace{-0.10in}
  \caption{Correlation of average generalization intervals with training and validation accuracy of using the models trained on the MNIST dataset using the actual labels.}
  \label{fig:mnist_cifar10_corr_class_5}
\end{figure}



As ReLU networks partition the input space
into linear regions~\citep{OnTheNumberofLinear14Montufar},
we consider two samples from same class denoted as $x_1$ and $x_2$.
The linear network can be defined as
\begin{equation}
    y_x = W_L(x)\ldots W_1(x) x,
    \label{eqy_x}
\end{equation}
where $x$ is the input and $W_1,\ldots, W_L$ are the weights of the layers 1,\ldots,L respectively. While the 
weights remain constant in one linear region, the weights in general depend on $x$, as indicated in Equation \ref{eqy_x}.

At a particular point, we have a specific linear network. Therefore, we have the following linear model at $x_1$. 
\begin{equation}\label{neqn1}
y_{x_1} = W_L(x_1)\ldots W_1(x_1)x_1
\end{equation}
If we consider a particular direction from  point $x_1$ to another point x, then we have the network as
\begin{equation}\label{neqn2}
y_{x} = W_L(x)\ldots W_1(x)(x_1+(x-x_1))
\end{equation}
The change of the output of the network in that particular direction is
\begin{equation}\label{neqn3}
\begin{aligned}
y_{x} - y_{x_1} = (W_L(x)\ldots W_1(x) - W_L(x_1)\ldots W_1(x_1))x_1 \\ + W_L(x)\ldots W_1(x)(x-x_1)
\end{aligned}
\end{equation}
If two training samples (i.e., $x_1$ and $x_2$) are classified correctly, then the difference between $y_{x_2}$ and $y_{x_1}$
should be small. If the two points in the same linear region, the first part of Equation \ref{neqn3} (i.e., $(W_L(x_2)\ldots W_1(x_2) - W_L(x_1)\ldots W_1(x_1))$) will be zero; if the linear regions are similar,
the part should be small. In general, we have established a bound of the change of weight matrices of ReLU networks;
see Appendix for detailed mathematical analysis. Currently we are working on improving the bound.
More generally, whether the change of the output (i.e., $y_{x} - y_{x_1}$) would be small or not along the direction ($x - x_1$) depends largely on the change of weight shown on the second part of the equation (i.e., $W_L(x)\ldots W_1(x)$). The part $(x-x_1)$ corresponds to tangent direction loosely when both $x_1$ and $x_2$ are on the same manifold. 




This observation is closely related to the one by \citeauthor{Tangent98Simard} [\citeyear{Tangent98Simard}], where learning should be more efficient along the tangent directions, compared to that of normal vectors on the underlying manifold. 
Their proposed tangent propagation algorithm makes sure that learning involves using the information of the derivatives of the classification function that is supplied by the tangent vectors. The tangent vectors are defined explicitly in their method, while in our case, we argue that the samples from the same class approximate the tangent vectors of the underlying manifold using pairwise interpolations defined by the data. For example, $(x_2-x_1)$ direction approximates a tangent vector on the underlying manifold, while $(x_1+x_2)$ approximates a normal direction.

The analysis shows that deep ReLU networks generalize via interpolation. 
In the random data point case, we have checked the generalization intervals that show that they generalize similarly like the real data points. If there is an underlying manifold, samples from the same class form a compact and possibly directional set, a small number of samples may be sufficient to approximate the manifold, enabling a trained network to generalize well from a small training set. When there is no manifold as in the random label case, it is difficult to characterize the set that a network generalizes to, but it still interpolates between samples with the same label.


\section{Experiments}

\subsection{Datasets}
Our experiments are based on the widely used representative MNIST ~\citep{Mnist1998Lecun} and CIFAR-10 datasets \citep{AlexCifar102009Learning}. The CIFAR-10 dataset consists of 32$\times$32  color images in 10 classes. The images are split into 50,000 training samples and 10,000 test/validation samples. On the other hand, The MNIST dataset consists of 70,000 handwritten digit images of size 28$\times$28 for each image; 60,000 training samples and 10,000 test/validation samples. However, to get some computational advantage, we have considered 1000 training samples (100 from each of the 10 classes). We have trained our deep neural network models with real labels and randomized labels after randomizing the class labels. We have made sure our network architectures and optimization procedures to get the training error to be very close to 0. 
We have mainly used the models that have three hidden dense layers with 128 ReLU units in each of the layers. However, to illustrate our proposed methods working on different architectures too, we have also conducted experiments on models with 1, 2, and 4 hidden layers having 512, 256, and 128 ReLU units respectively in each of the layers. 
The weights are initialized with random uniform, and the networks are optimized using SGD with learning rate of 0.01. The validation accuracy of the model trained on MNIST dataset is found around $90\%$ when trained on real labels (i.e., $0\%$ randomized) and $10\%$ when trained on random labels (i.e., $100\%$ randomized). To illustrate the impact of the dimension of input space on our proposed methods, we have also considered a two-dimensional synthetic dataset based on the paper by ~\citeauthor{Rozza2014Moon}~[\citeyear{Rozza2014Moon}]. This dataset is known as the ``two moons" dataset having an equal number of points in two classes (red and blue). Those points are split such that we get 300 training samples and 100 test/validation samples.

\subsection{Pairwise Interpolations That Approximate the Underlying Manifold}

Figure \ref{fig:tangent_normal} shows the output 
of the network trained on meaningful data (i.e., with real labels); along the path defined by samples from the same class, the neural network output for the correct class does not change much (i.e., stable) compared to other directions and the network classifies all the inputs along the path correctly. 
This behavior of the networks shows that they interpolate between the samples.

To demonstrate the neural networks' ability to approximate the underlying manifold,
we show the network outputs along the principal components for a class, which capture the tangent directions of the local manifolds. 
Similarly, minor components approximate normal directions. 
In a high-dimensional space, there are many normal directions to a local low-dimensional manifold. 
As examples, Figure \ref{fig:tangent_normal_pca}(left) shows the output along the first principal component of a particular class, while Figure \ref{fig:tangent_normal_pca}(middle) depicts the result along a normal direction by making the direction from class 9 (mean) to class 1 (mean) perpendicular to the first principal component. In addition, Figure \ref{fig:tangent_normal_pca}(right) shows the output for another component that has very small eigenvalue. 
Along the principal directions, the output of the correct class is more or less constant, which is expected as it approximates the tangent directions. 
Similarly, Figure \ref{fig:rl_same_random_class_pca_1st} illustrates the behavior of a
network trained on random labels. While the neural network output of the correct class changes more compared to that in
Figure \ref{fig:tangent_normal}(left), all the samples along the path are still classified
``correctly", i.e., the same as the random label assigned to the two samples. 
However, as expected, the mean of random class ``9" is not very different from that of any other random
class, and there is no meaningful local manifold to differentiate different classes.
Figure \ref{fig:rl_same_random_class_pca_1st}(right) verifies that there is indeed no manifold to approximate. 
However, the models still interpolate pairwise linearly as the other plots show. 



\subsection{Generalization Intervals for Lines and Triangles on MNIST and CIFAR-10}

To illustrate our idea of intrinsic generalization via interpolation, we have conducted a series of experiments on MNIST and CIFAR-10 datasets. 
For the pairwise lines, we observe almost similar generalization intervals for both real and random labels. 
In addition, to illustrate the role of the dimension of input space in interpolation, we explore generalization intervals on the MNIST dataset with reduced dimensions. 
To compute the generalization intervals, we consider the directions from particular training points to other points in the same class for both real and random labels. We have limited the maximal range as 3 (from -1 to 2) for pairwise lines centered at one of the two points. 
Figure \ref{fig:mnist_lines_class_5}(left) shows the distributions of generalization intervals with unchanged dimension of MNIST dataset ($28 \times 28$); the generalization intervals do not vary significantly between real and random labels.
As one expects, different paths may interfere with each other when they are close.
While the interference is small in the original dimension, the effects are shown as the small bump around 1.3.
When the dimension is reduced, the interference should be more prominent. 
 To demonstrate such effects, Figure \ref{fig:mnist_lines_class_5}(middle) and \ref{fig:mnist_lines_class_5}(right) illustrate the behavior when the samples are down-sampled. Both of the figures clearly show that, with lower dimensions, generalization intervals get lower with random labels. Note that there is no significant change in generalization intervals with real labels. This further confirms our claim of generalization via interpolation of the underlying manifold of the data. 
 We have also repeated the experiments on CIFAR-10 dataset. 
 As shown in Figure \ref{fig:gi_cifar10_al_rl},
 they behave similarly with real and random labels as on the MNIST dataset.

 
To further validate the effect of dimension of data, we observe the pairwise generalization intervals on 
``two moons" dataset. Figure \ref{fig:gi_2d_al_rl} illustrates the density of the generalization intervals for one of the two classes before and after randomizing the labels. This shows that the density changes dramatically after randomizing the labels of two-dimensional data. According to current definitions, apparently, it seems the network is memorizing data with random labels as the generalization intervals are much lower compared to data with real labels. 
In the two-dimensional case, almost all the lines have high probability of being interfered with each other. 

To study the generalization intervals of higher-order structures, Figure \ref{fig:mnist_triangles_class_5} shows the density of the generalization intervals along the directions defined by the pairwise lines connecting the midpoints of a triangle (i.e., the midlines; range as 5) formed by the samples of MNIST from class 5. We observe that the 
generalization intervals do not change much with those directions with real labels. However, we observe a significant change in the case of random labels. This confirms that there is no underlying manifold for random labels and
midline directions behave  like 
random directions.


We also find similar behavior with networks having different number of layers. Figure \ref{fig:mnist_lines_class_5_diff_layers}
shows the pairwise generalization interval distributions for three different architectures. Even though the number of layers is different, the distributions of the pairwise generalization intervals remain roughly the same. As the number of layers increases, the nonlinearity increases, and the bump
around 1.3 becomes more visible. 
Therefore, the experiments suggest
that generalization is largely decided by how the networks generalize between the points. In addition, the experiments confirm that the results are valid with different datasets and architectures. 

To further validate the relationship between generalization intervals and generalization performance of a neural network, Figure \ref{fig:mnist_cifar10_corr_class_5} shows the correlation of generalization intervals with training and validation performance for MNIST dataset during the training process. One can see clearly that
the generalization intervals correlate very well with
the model's generalization performance on the validation set.


\section{Discussion}
In this paper, by analyzing the mechanism of training deep neural networks,
we demonstrate quantitatively that deep neural networks generalize from training samples via interpolation, even when
the labels are randomized. The remarkable similarity of distributions of pairwise generalization intervals
on datasets with real and random labels answer the question that deep neural networks generalize and do not
simply memorize (i.e., associate a label to a particular input only). Furthermore, pairwise interpolations provide a good approximation of the underlying manifolds
of real datasets. 
The experiments lead to several further research questions.
For example, how regularization techniques and neural network architectures affect details
of specific solutions even though the pairwise generalization intervals are similar. 
In particular, the differences can and will affect the generalization performance
on particular validation sets as variations are expected. 
Additionally, as shown by \citeauthor{Zhang2016UnderstandingDL}~[\citeyear{Zhang2016UnderstandingDL}], the performance differences due to regularization techniques are often small. However, small differences could also be significant because models are often designed and trained for a small amount of improvements. It is important to analyze whether such improvements are extrinsic simply due to optimization or the choice of test sets or other factors. Similarly, as data are important in determining the generalization performance of deep neural networks, neural network architectures should be important. Clearly, different architectures can behave differently locally and therefore affect the details of generalization along pairwise paths. 
In this paper, we have used principal components to capture linear and local manifolds. 
Manifolds of real datasets are often complex and exhibit globally nonlinear structures. As the interpolations
are data-driven, we expect that the interpolations can also approximate nonlinear manifolds well.
How to quantify the relationships between nonlinearity of manifolds and deep neural network generalization needs to be further investigated. 


The correlation of generalization intervals with training and validation accuracy demonstrates
an intrinsic way to quantify the generalization performance of deep neural networks
without relying on particular choices of validation sets. The performance measure can be useful
for neural architecture search and for designing better neural network components and architectures. 
The linear interpolations, while effective for ReLU networks without any additional nonlinearity, need to
handle nonlinearity due to other nonlinear components such as max pooling and maxout. Such nonlinear components allow inputs not to be aligned but have the same outputs. For example, we have observed that the correlation between the generalization intervals and validation accuracy for the CIFAR-10 dataset is lower than that on the MNIST dataset. This is being investigated further.


\section{Conclusion}
In this paper, we demonstrate for the first time that deep ReLU neural networks
generalize through interpolations. While the pairwise generalization intervals
on real and random datasets are remarkably similar on high dimensional datasets such as
MNIST and CIFAR-10, the pairwise interpolations also approximate the underlying manifolds
well when they exist, enabling the models to generalize well.
While we have characterized systematically for networks with ReLU as the sole source of nonlinearity, how to compute the generalization
intervals efficiently for networks with additional nonlinearity such as max pooling and maxout still need to be investigated. Even though regularization techniques and neural network architectures may not have a significant impact on generalization, they can and do have impacts. Analyzing and modeling their impacts is also being studied.

\bibliographystyle{named}
\bibliography{main}
\newpage
\appendix
\section{Generalization and Interpolation of Deep ReLU 
networks based on Matrix Norm}

In this supplementary subsection, we apply basic operator theory to show that the rate of change of the output from the deep ReLU network is overall bounded with respect to the rate of change of the input. Moreover, this boundary is ``reasonable" in most portion of the input space. Then, we will further interpret the results shown in Figure 1, 2 and 3 of the main paper.\\

Let's start with deep linear networks. That is, we first remove the ReLU layers. A multi-layer linear network can be defined as
$$y = W_L\ldots W_1x,$$
where $x$ is the input and $W_1,\ldots, W_L$ are the matrices representation of layers 1,\ldots,L respectively. Then, suppose $W=W_L\ldots W_1$ and so that the linear network can be represented as $y=Wx$, which is a linear projection by matrix $W$ from $x\in\mathbb{R}^{n}$ to $y\in\mathbb{R}^{m}$.

Before going on, we shall briefly introduce the notions of vector's $p$-norm and matrix norm: Let $p\geq 1$ be a real number. Then, we can define a measure of distance called $p$-$\mathbf{norm}$ on the real space $\mathbb{R}^{n}$:
$$|\!|x|\!|_p := \left( \sum_{i=1}^n |x_i|^p \right)^{\frac{1}{p}}, \ \mathrm{for \ any} \  x=\{x_1,\ldots,x_n\}\in \mathbb{R}^n.$$

It only takes a routine derivation to show that $|\!|\cdot|\!|_p$ is indeed a norm, which is not discussed in this paper. However, we do want to note that: When $p=2$, we shall get the typical Euclidean distance. And when $p$ approaches positive infinity, the $p$-norm becomes the $\mathbf{infinite \ norm}$: 
$$|\!|x|\!|_{\infty}:= \max_{i=1,\cdots,n}|x_i|, \ \mathrm{for \ any} \  x=\{x_1,\ldots,x_n\}\in \mathbb{R}^n.$$ 

If we apply the same $p$-norm on both $\mathbb{R}^n$ and $\mathbb{R}^m$, there induces a corresponding $\mathbf{operator \ norm}$ (or $\mathbf{matrix \ norm}$) on the linear projection $y=Wx$ as:
$$|\!|W|\!|_p=\sup\left\{ \frac{|\!|Wx|\!|_p}{|\!|x|\!|_p} , \ \mathrm{for} \ x\in \mathbb{R}^n \ \mathrm{with} \ x\neq 0 \right\}.$$
where $\sup\{\}$ mean the supreme of the scales in this set. 

And we note that in the special case of $p=\infty$, the matrix norm will have the formula:
$$|\!|W|\!|_{\infty}=\max_{1\leq i\leq m}\sum_{j=1}^n|w_{ij}|,$$
which is simply the maximum absolute row sum of the matrix.

Intuitively, the matrix norm put an upper bound to the variance in $y$ with respect to that in $x$: $|\!|dy|\!|_p=|\!|Wdx|\!|_p\leq |\!|W|\!|_p\cdot|\!|dx|\!|_p.$ With this property, we provide a strict mathematical analysis on linear network generalization.\\

Suppose we are working with the dataset MNIST, so that $x\in\mathbb{R}^{784}$ and $y\in \mathbb{R}^{10}$, with each dimension $y_l$ in $y$ corresponding to the digit-class $l$. Also, we assume that the linear network $y=Wx$ is well-trained: Say, for any sample $x_i$, the dimension of the labeled digit-class in $y_i$ is larger than any other dimension by at least $\kappa=4.5$. In this case, suppose the dimension of the labeled digit-class is $L$. Then, $e^{{y_i}_L}/e^{{y_i}_l}=e^{{y_i}_L-{y_i}_l}\geq e^{4.5}\approx90$, and so that the output after softmax will be like $(0.01,\ldots,0.9,\ldots,0.01)$. But the exact value of $\kappa$ may be set differently based on the dimension of $y$ and the training requirements. We call $\kappa$ the $\mathbf{training \ requirement \ parameter}$. We shall put $\kappa$ in use shortly when interpreting experimental results.\\

Now, we add the ReLU layers back, so that
$$y=ReLU(W_L\cdot ReLU(W_{L-1}\cdots ReLU(W_1x)\cdots))$$

Suppose $y_1=W_1x$, and the $l$'th dimension of $y_1$ is less than zero. Then, the ReLU function will turn the $l$'th dimension of $y_1$ into zero, which is also equivalent to turning all the elements in row $l$ of $W_1$ into zeros. In fact, this means that for the fixed value $x$, we shall have $ReLU(W_1x)=W_1'x$, where $W_1'$ is the matrix same to $W_1$ except for some rows to be all zeros. 

We note that different values of $x$ will lead to different $W_1'$, i.e., different rows in $W_1$ will become zeros. But due to the continuity of $ReLU(W_1x)$, $W_1'$ will be fixed in local areas of input space $\mathbf{X}$. In fact, if $\dim(y_1)= m_1$, the input space $\mathbf{X}$ can be theoretically separated into $2^{m_1}$ regions: $\mathbf{X}=\cup_{r=1}^{2^{m_1}}\mathbf{X}_r$ with $\mathbf{X}_r\cap\mathbf{X}_s=\emptyset$ for $r\neq s$, so that $W_1'$ is fixed in each $\mathbf{X}_r$. Hence, we can see that the operation $ReLU(W_1x)$ is piece-wise linear throughout the input space $\mathbf{X}$. (Also note that we use $W_1(x)x$ to represent $ReLU(W_1x)$ in our main paper. But in this supplementary subsection, we always use $W_1'$ to represent the matrix $W_1$ after ReLU function.)

Taking this analysis into the entire deep ReLU network, we can have that: There exists a space separation $\mathbf{X}=\cup_{r=1}^{R}\mathbf{X}_r$ with $r<\infty$ and $\mathbf{X}_r\cap\mathbf{X}_s=\emptyset$ for any $r\neq s$, such that
\begin{align*}
y&=ReLU(W_L\cdot ReLU(W_{L-1}\cdots ReLU(W_1x)\cdots))\\
&=W_L'W_{L-1}'\cdots W_1'x=W'x,
\end{align*}
where each matrix $W_l'$ is the same to $W_l$ except for some rows to be all zeros, and all the matrices $W_1',\cdots,W_L'$ (and hence $W'$) are fixed in each $\mathbf{X}_r$. Therefore, we can see that the deep ReLU network is $\mathbf{piecewise \ linear}$ throughout the input space $\mathbf{X}$.

Since $W_1'$ is obtained by erasing rows in $W_1$ into zeros, it is obvious that $|\!|W_1'|\!|_{\infty}\leq |\!|W_1|\!|_{\infty}$ for any $W_1'$. However, although $|\!|W_2'W_1'|\!|_{\infty}\leq |\!|W_2W_1'|\!|_{\infty}$ holds true, $|\!|W_2W_1'|\!|_{\infty}\leq |\!|W_2W_1|\!|_{\infty}$ is not guaranteed. Hence, we cannot claim that $|\!|W_2'W_1'|\!|_{\infty}\leq |\!|W_2W_1|\!|_{\infty}$. Therefore, we cannot claim that $|\!|W'|\!|_{\infty}\leq |\!|W|\!|_{\infty}$.

But anyway, the deep ReLU network is indeed bounded by $\mathbf{K}=\sup_{r=1}^R\{|\!|W'|\!|_{\infty}, \ \mathrm{for \ each} \ W'$ $ \mathrm{corresponding \ to} \  \mathbf{X}_r\}$, due to $R<\infty$. In fact, we claim that for most $W'$, $|\!|W'|\!|_{\infty}$ is suppose to be in the same magnitude of $|\!|W|\!|_{\infty}$: Suppose $W'=W_2'W_1'$, i.e. $L=2$. The variance $|{W_2}_i\cdot {W_1'}_j|-|{W_2}_i\cdot {W_1}_j|$ should be randomly positive or negative. Then, if the row $i$ in $W_2'$ is not erased to zeros, we shall have ${W_2'}_i={W_2}_i$. And so that the difference between the absolute summation of $i$'th row in $W'$ and that in $W$ will be:
\begin{align*}
\sum_{j=1}^{n'}|{w'}_{ij}|-\sum_{j=1}^{n'}|{w}_{ij}|&=\sum_{j=1}^{n'}|{W_2}_i\cdot {W_1'}_j|-\sum_{j=1}^{n'}|{W_2}_i\cdot {W_1}_j|\\
&=\sum_{j=1}^{n'}\left(|{W_2}_i\cdot {W_1'}_j|-|{W_2}_i\cdot {W_1}_j|\right).
\end{align*}
According to this form, the absolute summation in one row of $W'$ shall hardly show significant change comparing to the same row in $W$. This is because most amount of variance may be cancelled out due to the randomness in $|{W_2}_i\cdot {W_1'}_j|-|{W_2}_i\cdot {W_1}_j|$. Hence, $|\!|W'|\!|_{\infty}\approx|\!|W|\!|_{\infty}$ shall hold true for most $W'$ when $L=2$. When $L\geq 3$, we may set $W''=W_{L-1}'\cdots W_1'$, so that $W'=W_L'\cdot W''$. Then the same result shall be obtained by mathematical induction.

We use the concept ``most" for many times in the above analysis. In fact, the more nodes in the hidden layers, the closer $|\!|W'|\!|_{\infty}$ and $|\!|W|\!|_{\infty}$ should be. This is because the  random differences caused by ReLU may be balanced in the high dimension of row ${W_{l+1}}_i$ and column ${W_{l}'}_j$. Given that our deep ReLU network has 3 hidden layers with 128 nodes in each layer, we assume that the probability of any input $x$ to locate in a region $\mathbf{X}_r$ with $|\!|W'|\!|_{\infty}\leq 2 |\!|W|\!|_{\infty} $ is 99$\%$. We call this statement the $\mathbf{norm \ distribution \ assumption}$.\\

Now, we are ready to interpret the experimental results in the previous section. Once again, we only use MNIST as our example. The situations with CIFAR-10 and other datasets shall be the same.

 We first deal with the performance of deep ReLU network trained by actual labels. We assume that the training samples are ``dense", with samples from the same class clustered together. Actually, dense and clustered samples are typical for many datasets. And these clusters usually form specific structures such as manifolds or hyper-planes. But to be specific, we assume that there is a domain $\mathbf{S}$ in the sample space satisfying:

$(i)$ Every sample in $\mathbf{S}$ belongs to the class $L$. 

$(ii)$ For any $x\in \mathbf{S}$, there exists a sample $x_i$ from class $L$ such that $|\!|x-x_i|\!|_{\infty}< \frac{\kappa}{4|\!|W|\!|_{\infty}}.$ Given the MNIST dataset, this means that each pixel in $x$ is within the range of $\frac{\kappa}{4|\!|W|\!|_{\infty}}$ of the corresponding pixel in $x_i$.

Then, for any input $x\in \mathbf{S}$, suppose $x_i$ is its closest sample. We assume that $x_i\in\mathbf{X}_{r_0}$ and $x\in \mathbf{X}_{r_1}$, where $\mathbf{X}_{r_0}$ and $\mathbf{X}_{r_1}$ are the small regions with fixed $W'$, corresponding to the well-trained deep ReLU network $W$. Then, suppose the line segment $\overrightarrow{x-x_i}$ (starting at $x_i$ and ending at $x$) passes through the $R$ boundaries between $R+1$ small regions $\mathbf{X}_{r_0}=\mathbf{X}_0,\cdots,\mathbf{X}_{R}=\mathbf{X}_{r_1}$. Then, suppose the points $\hat{x}_1,\cdots,\hat{x}_R$ are on each boundary respectively. 

According to the definition of $\mathbf{S}$, we have that $|\!|x-x_i|\!|_{\infty}<\frac{\kappa}{4|\!|W|\!|_{\infty}}$, and there exists positive numbers $\alpha_0,\alpha_1,\cdots,\alpha_R$ with $\sum_{r=0}^R\alpha_r=1$ such that $|\!|\hat{x}_{r+1}-\hat{x}_{r}|\!|_{\infty}=\alpha_r\cdot|\!|x-x_i|\!|_{\infty}$ (we define $\hat{x}_0=x_i$ and $\hat{x}_{R+1}=x$). In addition, since the deep ReLU network $W$ is continuous, inputs on the boundary between two regions can take either $W'$ from each region. That is, $W_{r-1}'\hat{x}_r=W_{r}'\hat{x}_r$ for $r=1,2,\cdots,R$. 

Then, we have (for simplicity, we use $|\!|\cdot|\!|$ to note $|\!|\cdot|\!|_{\infty}$)
\begin{align*}
&|\!|y-y_i|\!|=|\!|W_R'x-W_0'x_i|\!|\\
&=\!|\!|W_R'\hat{x}\!_{R+1}\!-\!W_R'\hat{x}\!_{R}\!+\!W_{R-1}'\hat{x}\!_{R}\!-\!\cdots\!\!-\!\!W_1'\hat{x}_1\!\!+\!\!W_0'\hat{x}_1\!\!-\!\!W_0'\hat{x}_0|\!|\\
&\leq\!\!|\!|\!W_R'\hat{x}\!_{R\!+\!1}\!\!-\!\!W_R'\hat{x}\!_{R}|\!|\!\!+\!\!|\!|\!W_{R\!-\!1}'\!\hat{x}\!_{R}\!\!-\!\!W_{R\!-\!1}'\!\hat{x}\!_{R\!-\!1}\!|\!|\!\!+\!\!\cdots\!\!+\!\!|\!|\!W_0'\hat{x}_1\!\!-\!\!W_0'\hat{x}_0\!|\!|\\
&=\!|\!|W_R'(\hat{x}_{R+1}\!\!-\!\!\hat{x}_{R})|\!|\!+\!|\!|W_{R\!-\!1}'\!(\hat{x}_{R}\!\!-\!\!\hat{x}_{R\!-\!1})\!|\!|\!+\!\cdots\!+\!|\!|W_0'(\hat{x}_1\!\!-\!\hat{x}_0\!)\!|\!|\\
&\leq |\!|W_R'|\!||\!|\hat{x}_{R+1}\!\!-\!\!\hat{x}_{R}|\!|\!\!+\!\!|\!|W_{R\!-\!1}'|\!||\!|\hat{x}_{R}\!\!-\!\!\hat{x}_{R\!-\!1}\!|\!|\!+\!\cdots\!+\!|\!|W_0'|\!||\!|\hat{x}_1\!\!-\!\hat{x}_0\!|\!|
\end{align*}

According to our norm distribution assumption, 99$\%$ of $W_r'$ will have $|\!|W_r'|\!|\leq2|\!|W|\!|$. We assume that $|\!|x-x_i|\!|_{\infty}<\frac{\kappa}{4|\!|W|\!|_{\infty}}$, which is a small range. So, the value $R$ should not be huge enough to have more than 2 or 3 $W_r'$ with $|\!|W_r'|\!|>2|\!|W|\!|$. As a result, we can say that in most cases, we have
\begin{align*}
|\!|y-y_i|\!|&\leq 2|\!|W|\!| \left(|\!|\hat{x}_{R+1}\!\!-\!\!\hat{x}_{R}|\!|\!+\!|\!|\hat{x}_{R}\!\!-\!\!\hat{x}_{R\!-\!1}\!|\!|\!+\!\cdots\!+\!|\!|\hat{x}_1\!\!-\!\hat{x}_0|\!|\right)\\
& =\! 2|\!|W|\!|\!(\alpha_R|\!|x\!-\!x_i|\!|\!+\!\alpha_{R\!-\!1}\!|\!|x\!-\!x_i|\!|\!+\cdots \alpha_0|\!|x\!-\!x_i|\!|)\\
&=2|\!|W|\!|\cdot|\!|x-x_i|\!|<2|\!|W|\!|\cdot \frac{\kappa}{4|\!|W|\!|}=\frac{\kappa}{2}
\end{align*}

This means that each dimension in $y$ is within the range of $\frac{\kappa}{2}$ of the corresponding dimension in $y_i$. So for any dimension $y_l$ in $y$, we have that:
\begin{align*}
y_L-y_l&=(y_L-{y_i}_L)+{y_i}_L-(y_l-{y_i}_l)-{y_i}_l\\
&=y_L\!-\!y_l=[(y_L\!-\!{y_i}_L)\!-\!(y_l\!-\!{y_i}_l)]\!+\!({y_i}_L\!-\!{y_i}_l)\\
&>[(-\frac{\kappa}{2})-(\frac{\kappa}{2})]+(\kappa)=0
\end{align*}
That is, the dimension of class $L$ in $y$ will not be surpassed by any other dimension, which means that $x$ shall be classified into $L$. Finally, due to the ambiguity of $x$ in $\mathbf{S}$ and the norm distribution assumption, we can conclude that the deep ReLU network $W$ is correctly generalized to most portion of the domain $\mathbf{S}$.

This conclusion can explain the results shown by Figure 1 and 2 well: As we said, most dense cluster domains $\mathbf{S}$ preserve specific structures such as manifolds or hyper-planes in the input space. Hence, when moving along two samples $x_1$, $x_2$ from the same class, or along the first principle component direction, it is very likely that $\overrightarrow{x_2-x_1}$ is within the manifold $\mathbf{S}$. Hence, the classification won't change due to the generalization of the deep ReLU network in $\mathbf{S}$. 

However, when $x_1$ and $x_2$ come from different classes, or when  $x$ moves perpendicular to the first principle component along normal directions, the most common situation is that $\overrightarrow{x_2-x_1}$ will move from domain $\mathbf{S}_1$ to domain $\mathbf{S}_2$, so that the classification will change with steady when x passes through the boundary of the two domains. But when moving from $x_1$ along a random direction, it is very likely that $\overrightarrow{x-x_1}$ will go beyond the hyperplane and arrive in areas where samples are sparse. Hence, the deep ReLU network are not well generalized in these areas and dimensions in $y$ shall oscillate with chaos.

But when we randomly assign the labels, the manifolds with dense clustered samples will be broken down into pieces according to samples assigned with different labels. So, there is hardly any dense domain $\mathbf{S}$ for the deep ReLU network to produce a good generalization. As a result, moving along any directions are less stable than the case with actual labels, as shown in Figure 3. Specifically, moving along the first principle component direction has no difference than moving along any random direction. This is because the random labels break down each manifold $\mathbf{S}$, so that no principle component direction exists anymore.

However, even with random labels, if both $x_1$ and $x_2$ are from the same class, moving along $\overrightarrow{x_2-x_1}$ is relatively stable. This is because the deep ReLU network is still piece-wise linear and piece-wise bounded by $|\!|W'|\!|$ along $\overrightarrow{x_2-x_1}$. So if $y_1$ and $y_2$ are almost the same, then all the dimensions in $y$ cannot oscillate ``at will" during the moving along $\overrightarrow{x_2-x_1}$, since at last each dimension in $y$ need to go back to the same. Hence, even with actual labels, moving along $\overrightarrow{x_2-x_1}$ with the same label on $x_1$ and $x_2$ are relatively stable.\\

In summary, we show that a deep ReLU network does piece-wise linear interpolation rather than memorization in the input space. And its interpolation is bounded mostly. Hence, if samples from the same classes form dense clusters in the input space, the deep ReLU network shall generalize well in the domains covered by the dense clusters.

\end{document}